\documentclass[pdflatex,sn-basic]{sn-jnl_adjusted} 
\usepackage{amsmath}
\usepackage{bm}

\theoremstyle{thmstylethree}%
\newtheorem{dfn}{Definition}

\newcommand{\eone}{E_{100}}
\newcommand{\etwo}{E_{1000}}
\newcommand{\ethree}{U_3}
\newcommand{\efour}{U_{1003}}
\newcommand{\nclust}{n_c}
\newcommand{\kumap}{k}
\newcommand{\nobs}{n_{\operatorname{obs}}}

\begin{document}

\title[Enhancing cluster analysis via topological manifold learning]{Enhancing cluster analysis via topological manifold learning}

\author*[1]{\fnm{Moritz} \sur{Herrmann}}\email{moritz.herrmann@stat.uni-muenchen.de}

\author[2]{\fnm{Daniyal} \sur{Kazempour}}\email{dka@informatik.uni-kiel.de}

\author[1]{\fnm{Fabian} \sur{Scheipl}}\email{fabian.scheipl@stat.uni-muenchen.de}
\equalcont{These authors contributed equally to this work.}

\author[2]{\fnm{Peer} \sur{Kröger}}\email{pkr@informatik.uni-kiel.de}
\equalcont{These authors contributed equally to this work.}

\affil[1]{\orgdiv{Department of Statistics}, \orgname{Ludwig-Maximilians-Universität München}, \orgaddress{\street{Ludwigstr. 33}, \city{Munich}, \postcode{80539}, \state{Bavaria}, \country{Germany}}}

\affil[2]{\orgdiv{Department of Computer Science}, \orgname{Christian-Albrechts-Universität zu Kiel}, \orgaddress{\street{Christian-Albrechts-Platz 4}, \city{Kiel}, \postcode{24098}, \state{Schleswig-Holstein}, \country{Germany}}}

\abstract{
We discuss topological aspects of cluster analysis and show that inferring the topological structure of a dataset before clustering it can considerably enhance cluster detection: theoretical arguments and empirical evidence show that clustering embedding vectors, representing the structure of a data manifold instead of the observed feature vectors themselves, is highly beneficial. To demonstrate, we combine manifold learning method UMAP for inferring the topological structure with density-based clustering method DBSCAN. Synthetic and real data results show that this both simplifies and improves clustering in a diverse set of low- and high-dimensional problems including clusters of varying density and/or entangled shapes. Our approach simplifies clustering because topological pre-processing consistently reduces parameter sensitivity of DBSCAN. Clustering the resulting embeddings with DBSCAN can then even outperform complex methods such as SPECTACL and ClusterGAN. Finally, our investigation suggests that the crucial issue in clustering does not appear to be the nominal dimension of the data or how many irrelevant features it contains, but rather how \textit{separable} the clusters are in the ambient observation space they are embedded in, which is usually the (high-dimensional) Euclidean space defined by the features of the data. Our approach is successful because we perform the cluster analysis after projecting the data into a more suitable space that is optimized for separability, in some sense.
}

\keywords{Cluster analysis, Manifold learning, Topological data analysis}

\maketitle

\section{Introduction}
Clustering is the task of uniting similar and separating dissimilar observations in a dataset \citep{kriegel2009clustering, aggarwal2014introduction}. It is a fundamental task in data analysis and is thus widely investigated in many fields. With this study, we intend to raise awareness for topological aspects of clustering and to provide empirical evidence that topologically-informed approaches which are conceptually and computationally simple can compete with or even outperform much more complex existing methods on a wide range of problems.

\subsection{Problem specification}
Cluster analysis is usually approached in an algorithm-driven manner, and considerations about the underlying principles of data generating processes and data structures are often limited to a probabilistic conceptualization assuming that the data $X$ follow a joint probability distribution $P(X)$ \citep{hastie09esl} or, more precisely, a mixture of distributions \citep{aggarwal2014introduction}. In contrast, connections to topological data analysis (TDA) \citep{chazal2017introduction, wasserman2018topological}, a branch of statistical data analysis inferring the structure of data leveraging topological concepts, are usually not considered. In general, the topological aspects of cluster analysis appear to be an under-investigated topic. Current textbooks on cluster analysis \cite[e.g]{aggarwal2014data, aggarwal2015data, giordani2020introduction, scitovski2021cluster, hennig2015handbook} and recent reviews of the field \cite[e.g.]{jain1999data, kriegel2009clustering, assent2012clustering, pandove2018systematic, mittal2019clustering}  rarely mention the term ``topology". 

Following \cite{niyogi2011topological}, we consider clustering a natural example of TDA. Since an improved understanding of the underlying principles governing the problem is likely to lead to more suitable methods and novel solutions, our work aims to reduce this lack of awareness of topological aspects in the clustering literature. Specifically, our approach follows \citet[p. 2]{niyogi2011topological} who state that \lq\lq clustering is a kind of topological question" which tries to separate the data into \lq\lq connected components." One particularly relevant consequence of this topological perspective is its implication that the difficulty of a clustering problem is not necessarily determined by the data's (nominal) dimensionality.

\subsection{Scope of the study}
 
In this work, we make use of the well-known algorithm DBSCAN \citep{dbscan} for cluster detection and the recently developed manifold learning algorithm UMAP \citep{mcinnes2018umap} to infer the topological structure of a dataset. 

UMAP has a decidedly topological underpinning, so it is suitable for a theoretical analysis from the clustering perspective we take here. In particular, it builds on simplicial complexes to obtain a fuzzy topological representation of the inherent structure of a dataset. As such, it is based on the same theoretical principles as topological data analysis \citep{chazal2017introduction, wasserman2018topological}. 
In addition, it has already been shown that preprocessing by UMAP can improve clustering results \citep{allaoui2020considerably} and that the resulting embeddings frequently yield ``more compact clusters than t-SNE [another state-of-the-art manifold learning method] with more white space in between" \cite[p. 157]{kobak2021initialization}.

To be specific, \lq\lq inferring the topological structure" as we do here with UMAP has two aspects: first, a fuzzy graph representation of the dataset is used to find the (number of) connected components. Second, this structure is represented by embedding vectors (i.e. coordinates in a representation space) that are optimized for the separability of the connected components. As we show in section \ref{sec:problems}, UMAP's graph construction and graph embedding steps both increase cluster separability, and their combined effect thus improves clusterability dramatically.

DBSCAN, on the other hand, is a widely used and well-established method for cluster detection \citep{schubert2017dbscan}. In particular, it neither requires a pre-specified number of clusters nor does it make any assumptions about their specific shapes or patterns. This is important, as inferring the connected components of a dataset is largely equivalent to identifying the clusters it contains. Moreover, the optimized representation of the topological approach focuses on the separability of clusters, not on the specific shapes the clusters might have. Also note that UMAP's developers conjectured that it might enhance density-based clustering, but that this requires further investigation \citep{umapdoc}.

From a practical perspective, this means we use UMAP to preprocess the data such that its representation is optimized for separability and use the resulting embedding vectors as inputs for DBSCAN. Although the theoretical and empirical considerations outlined above show that these two methods are suitable, it has to be emphasized that this does not mean that we consider UMAP and DBSCAN the most suitable combination in general. Certainly, additional research has to focus on the pros, cons, and differences between UMAP and other manifold learning methods, in particular t-SNE, and some efforts have already been made in this direction \citep{kobak2021initialization, wang2021understanding}. In this paper, we intend to show that a topological perspective, in general, can improve understanding and practical feasibility of clustering and not whether that specific combination of methods is the most suitable. Other combinations of clustering and/or manifold learning methods than UMAP and DBSCAN are possible and certainly deserve investigation as well. 

Moreover, note that there are other approaches to infer the topological structure of a dataset. For example, persistent homology -- which also builds on simplicial complexes -- quantifies the topological structure of a dataset by providing information on statistically significant persistent topological features such as connected components, holes, or voids, e.g. \citep{wasserman2018topological}. In contrast, measures of data separability such as the distance-based separability index \citep{guan2021novel} quantify the separability of datasets in a single scalar value. However, both approaches only contribute to the first aspect of inferring the topological structure, i.e. they do not provide data representation optimized for separability. 

\subsection{Contributions}

This study makes three distinct contributions: 
First, section \ref{sec:problems} illustrates that approaches motivated by a topological perspective can dramatically reduce the complexity of clustering for both low- and high-dimensional data. This is achieved with an in-depth analysis of simulated data specifically designed to reflect some often described problems of clustering including high-dimensional data, clusters of different densities, and irrelevant features. In addition, a simple toy example demonstrates why and how inferring the intrinsic topological structure of a dataset with UMAP before clustering improves the clustering performance of DBSCAN. 

Secondly, with intuition and motivation in place, section \ref{sec:price} is devoted to specific implications of the topological perspective. We describe which structures of a dataset are preserved when inferring the topological structure by finding connected components and enhancing separability (using the UMAP algorithm), in particular by contrasting topological against geometrical characteristics in a detailed qualitative and quantitative analysis of simple synthetic examples.

Finally, in section \ref{sec:experiments}, we report extensive experiments using real-world data. Our results show that inferring the topological structure of datasets before clustering them not only improves -- dramatically, for some examples such as MNIST -- performance of DBSCAN, but also drastically reduces its parameter sensitivity. The comparatively simple approach of combining UMAP and DBSCAN can even outperform recently proposed clustering methods such as ClusterGAN \citep{mukherjee2019clustergan}, which require expensive hyperparameter tuning, on complex datasets.  

In addition, related work and the methods used are described in section \ref{sec:background}, while the results are discussed in section \ref{sec:discussion} before we conclude in section \ref{sec:conclusion}.

\section{Methods and related work}\label{sec:background}

In this section, we first describe the background of the study and related work, before we outline the methods DBSCAN and UMAP, which are used for clustering and inferring topological structure, respectively, in this study. Readers which are familiar with the methods might skip the corresponding paragraphs. However, note that we will refer to some of the more technical details outlined here in section \ref{sec:problems:toy}.

\subsection{Background and related work}\label{sec:related-work}

The body of literature on clustering, topological data analysis, and manifold learning is extensive and has seen contributions from many different areas and perspectives. General reviews on clustering have been provided for example by \cite{jain1999data} and more recently by \cite{saxena2017review}. Moreover, there a several reviews focusing on cluster analysis for high-dimensional data \citep{kriegel2009clustering, assent2012clustering, pandove2018systematic, mittal2019clustering}. In addition, there exist overviews on TDA \cite[e.g.]{niyogi2011topological, chazal2017introduction, wasserman2018topological} as well as on manifold and representation learning \citep{cayton2005algorithms, bengio2013representation, wang2021understanding} including the textbooks by \cite{ma2012manifold} and \cite{lee2007nonlinear}.

The variety of clustering algorithms is vast and endeavors have been made to capture this diversity through taxonomies. DBSCAN, the algorithm used here, is a density-based approach. One of its major advantages is that it does not require a pre-specified number of clusters and that the clusters can have arbitrary shapes and patterns. Its hierarchical version \citep[HDBSCAN,][]{hdbscan} does not use a global $\varepsilon$-threshold but computes on its own multiple cut-off values resulting in clusters of different densities and therefore requires only the $minPts$ parameter. Similar to HDBSCAN, the OPTICS algorithm \citep{optics} calculates an ordering of the observations without a global $\varepsilon$-threshold that provides broader insight on the structure of the data. However, the method does not explicitly assign cluster memberships. Instead, it allows viualizing the hierarchical cluster structure for example via reachability plots \citep{optics}.

Further categories are \textit{hierarchical} and \textit{partitioning} algorithms \citep{jain1999data}, where the latter can be divided further into sub-taxonomies. Some of them are based on the minimization of distances to certain prototypes (centroids, medoids, etc.), this includes algorithms like $k$-means \citep{kmeans}, or its more general archetype of algorithms: Gaussian Mixture Models (GMMs) among which the Expectation-Maximization (EM) algorithm \citep{emalgo} is a prominent exponent. A major caveat, however, is that these methods estimate a specific probabilistic model which includes the number of clusters to be detected and often fail if the data is distributed differently \citep{liu2014spectral}.

In contrast, \textit{spectral} clustering, a family of algorithms that shares some common ground with many manifold learning methods that are also based on spectral decompositions of pairwise (dis)similarity matrices, is more robust with respect to the shape and distribution of the clusters. However, these methods require the number of clusters to be specified in advance \citep{luxburg2007tutorial, liu2014spectral}. 

\textit{Subspace clustering} approaches emerged specifically for high-dimensional settings  \citep{kriegel2009clustering, assent2012clustering, pandove2018systematic, mittal2019clustering}. 
The fundamental assumption here is that objects within a cluster do not exhibit high similarities among all dimensions but only within a small subset of features that can either (a) span an \textit{axis-parallel} subspace or (b) an affine projection to an  \textit{arbitrarily-oriented} subspace (``correlation clustering''). In both cases, the objects of a cluster are assumed to be located on a common, low-dimensional linear manifold. 

In contrast, manifold learning is based on the assumption that data observed in a high-dimensional ambient observation space is distributed on or near a potentially nonlinear manifold with a much smaller intrinsic dimension than the ambient space \citep{ma2012manifold}. 
In general, the aim is to find low-dimensional representations of datasets preserving as much of the structure of the observed data as possible. A synonymous term is nonlinear dimension reduction (NDR) \citep{lee2007nonlinear}. However, there is no general definition of which characteristics are to be preserved and represented and different methods infer the intrinsic structure and provide low-dimensional representations in different ways. 

For instance, principal component analysis (PCA) yields embedding vectors that optimally preserve global Euclidean distances in the original data space, while other methods such as Isomap \citep{tenenbaum2000global} yield embedding vectors that aim to preserve geodesic distances on a single, globally connected data manifold. Methods like t-distributed Stochastic Neighbor Embedding \citep[t-SNE,][]{van2008visualizing} and  uniform manifold approximation and projection  \citep[UMAP,][]{mcinnes2018umap} have been successfully applied to complex high-dimensional datasets with cluster structure. More recently, methods with a specific topological focus such as general purpose Topomap \citep{doraiswamy2020topomap} as well as domain specific Paga \citep{wolf2019paga}, which focuses on the analysis of single cell data, have been proposed. The manifold learning-based clustering approach of \cite{souvenir2005manifold} relies on the assumption that data is sampled from multiple \textit{intersecting} lower-dimensional manifolds.

Several studies that precede ours also focus on the combination of manifold learning techniques and cluster analysis, with applications to cytometry data \citep{putri2019dimensionality}, brain tumor segmentation \citep{kaya2017pca}, spectral clustering \citep{arias2017spectral}, or big data \citep{feldman2020turning}, the latter three based on PCA. DBSCAN was used in combination with multi-dimensional-scaling (MDS) in \cite{mu2020study}, and UMAP was used for time-series clustering \citep{pealat2021improved} as well as clustering SARS-COV-2 mutation datasets \citep{hozumi2021umap}. However, these all focus on specific domains and not on the underlying topological principles. 
In contrast, we base our work on a topological perspective on clustering first described theoretically by \cite{niyogi2011topological}, who conceptualize clustering as the problem of identifying the \emph{connected components} of a data manifold. We show the theoretical and practical utility of this perspective by means of extensive experiments based on synthetic and real datasets. 
Similar in spirit to our work, \cite{allaoui2020considerably} perform a comparative study with real data to show that UMAP can considerably improve the performance of clustering algorithms. Among other things, they combined UMAP with HDBSCAN and report comparable clustering results for three of the real-world datasets (Pendigits, MNIST and FMNIST) also used here. 
However, in contrast to our study, \cite{allaoui2020considerably} do not provide insights into the conceptual topological underpinnings, nor do they describe how the data structures preserved in UMAP embeddings lead to these performance improvements. Note that their results also show empirically that the benefits of the proposed approach are not tied to any particular combination of NDR and clustering methods.

\subsection{UMAP} 

The principle idea behind UMAP essentially consists of two steps: \\
\textit{1)} Constructing a weighted $\kumap$-nearest neighbor ($\kumap$-NN) graph from a pairwise distance matrix.\\
\textit{2)} Finding a (low-dimensional) representation of the graph which preserves as much of its structure as possible. \\
Note that this is the fundamental principle in manifold learning and the details of the two steps constitute the differences between manifold learning methods \citep{wang2021understanding}. However, unlike many other manifold learning methods, UMAP is based on a solid theoretical foundation that ensures that the topology of the manifold is faithfully approximated by its fuzzy simplical set representation. We concentrate on the computational aspects outlined in \cite{mcinnes2018umap} and refer interested readers to the original study for theoretical details. 

\subsubsection{Graph construction}

Given a dataset $X = \{x_1, ..., x_{\nobs}\}$ sampled from a space equipped with a distance metric $d(x_i, x_j)$, UMAP constructs a directed $\kumap$-NN graph $\Bar{G} = (V, E, w)$ with the vertices $V_i$ being observations $x_i$ from $X$, $E$ the edges and $w$ the weights, based on the following definitions.

\begin{dfn}\label{def:umap1}
The distance $\rho_i$ of an observation $x_i$ to its nearest neighbor $x_{i_j}$ is defined by 
$$\rho_i = \min\{d(x_i, x_{i_j}) \vert 1 \leq j \leq k, d(x_i, x_{i_j}) > 0\}.$$
\end{dfn}

\begin{dfn}\label{def:umap2}
A (smooth) normalization factor $\sigma_i$ is set for each $x_i$ by
$$\sum^k_{j=1}\exp\left(\frac{-\max(0, d(x_i, x_{i_j}) - \rho_i)}{\sigma_i}\right) = \log_2(k).$$
This defines a local (Riemannian) metric at point $x_i$.
\end{dfn}

\begin{dfn}{Weight function:}\label{def:umap3}
The edge weights of the graph are defined by 
$$w((x_i, x_{i_j})) = \exp\left(\frac{-\max(0, d(x_i, x_{i_j}) - \rho_i)}{\sigma_i}\right).$$ 
Note, the distance to the nearest neighbor $\rho_i$ ensures that $x_i$ is connected to at least one other point with an edge of weight 1 (local connectivity constraint).
\end{dfn}

For the theory to work it is essential to assume that the data is uniformly distributed on the manifold, which is too strong an assumption for real-world data. The issue is bypassed by defining independent notions of distance at each observed point through $\sigma_i$ and $\rho_i$. However,  these local metrics may not be mutually interchangeable, which means that the ``distance'' between neighboring points $x_i$ and $x_j$ may not be the same if measured w.r.t $x_i$ or w.r.t. $x_j$, i.e., $d(x_i, x_j) \neq d(x_i, x_j)$, so edge weights in $\Bar{G}$ depend on the direction of the edges. 

A unified, undirected graph $G$ with adjacency matrix $B$ is obtained by
\begin{equation}
    B = A + A^T - A \circ A^T, 
    \label{eq:had}
\end{equation}
 with $A$ the weighted adjacency matrix of $\Bar{G}$ and $\circ$ the point-wise product. Note that Eq. \eqref{eq:had} represents the well-defined operation of unioning fuzzy simplicial sets (with which the manifold is approximated). The resulting entries in $B$ can be interpreted as the probability that at least one of the two directed edges between two vertices in $\Bar{G}$ exists, or more generally as a measure of similarity between two observations $x_i$ and $x_j$. Note that it has recently been shown that a stricter notion of connectivity induced by mutual nearest neighbors can further improve the topology preserving property of standard UMAP used here \citep{dalmia2021clustering}. 

\subsubsection{Graph embedding}

The objective is to find a configuration of points in the representation space $Y$ whose fuzzy simplicial set is as similar as possible to the fuzzy simplicial set of the original data, as represented by $G$. 
To find this low-dimensional representation, UMAP optimizes the cross entropy of edge weights in the two spaces.  Similarities in the observation space are represented in terms of the local smooth nearest neighbor distances as
\begin{equation}\label{eq:vij}
    v_{ij} = (v_{j \vert i} + v_{i \vert j}) - v_{j \vert i}v_{i \vert j}, 
\end{equation}
with $v_{j \vert i} = \exp[(-d(x_i, x_j) - \rho_i)/\sigma_i]$ (c.f. Eq. \eqref{eq:had}), and similarities in the representation space $Y$ as
\begin{equation}\label{eq:wij}
    w_{ij} = (1 + a \vert\vert y_i - y_j \vert\vert ^{2b}_2)^{-1},
\end{equation}

the cross entropy between the two fuzzy simplicial set representations 

\begin{equation}\label{eq:crosse}
    C_{UMAP} = \sum_{i \neq j} v_{ij}\log\left(\frac{v_{ij}}{w_{ij}}\right) + (1 - v_{ij})\log\left(\frac{1 - v_{ij}}{1 - w_{ij}}\right) 
\end{equation}

is minimized via stochastic gradient descent (SGD) to obtain the graph layout (by default $a \approx 1.929$ and $b \approx 0.7915$). The two terms in Eq. \eqref{eq:crosse} represent the attractive and repulsive forces for the graph layout algorithm used here.\\
Next to $a$ and $b$, UMAP's central tuning parameters are the number of nearest neighbors $\kumap$ (often denoted as $n$ or \verb|n_neighbors|), the number of SGD optimisation iterations $\verb|n-epochs|$, the dimension $d$ of the representation space, and $\verb|min-dist|$, a parameter controlling how close neighboring points can appear in the representation.

\subsection{DBSCAN}

The principle idea behind DBSCAN is captured within 6 definitions we adapt from \cite{dbscan} and elaborate on:

\begin{dfn}{$\varepsilon$-neighborhood of an object:}\label{def:db1}
The $\varepsilon$-neighborhood of an object $x_i$ denoted by $\mathcal{N}_{\varepsilon}(x_i)$, is defined by:
$$ \mathcal{N}_{\varepsilon}(x_i) = \lbrace x_j \in X \vert d(x_i,x_j) \leq \varepsilon \rbrace $$
where $X$ denotes a given dataset. 
\end{dfn}

\begin{dfn}{Directly density-reachable:}\label{def:db2}
An object $x_i$ is direct density-reachable from an object $x_j$ w.r.t. a given $\varepsilon$-range and $MinPts$ if:\\
1) $x_i \in \mathcal{N}_{\varepsilon}(x_j)$ and\\
2) $\vert \mathcal{N}_{\varepsilon}(x_j) \vert \geq MinPts$ (core point condition)
\end{dfn}

\begin{dfn}{Density-reachable:}\label{def:db3}
An object $x_i$ is density-reachable from another object $x_j$ w.r.t. $\varepsilon$ and $MinPts$ if there is a chain of objects $x_1,...,x_c$, $x_1 = x_i$, $x_c = x_j$ such that $x_{l+1}$ is directly density-reachable from $x_l$.
\end{dfn}

\begin{dfn}{Density-connected:}\label{def:db4}
An object $x_i$ is density-connected to another object $x_j$ w.r.t. $\varepsilon$ and $MinPts$ if there is an object $o$ such that both, $x_i$ and $x_j$ are density-reachable from $o$ w.r.t. $\varepsilon$ and $MinPts$.
\end{dfn}

\begin{dfn}{Cluster:}\label{def:db5}
Let $X$ be a given dataset of objects. A cluster $C$ w.r.t. $\varepsilon$ and $MinPts$ is a non-empty subset of $X$ satisfying the following conditions:\\
1) $\forall x_i,x_j:$ if $x_i\in C$ and $x_j$ is density-reachable from $x_i$ w.r.t. $\varepsilon$ and $MinPts$, then $x_j \in C$ (Maximality)\\
2) $\forall x_i,x_j \in C: x_i$ is density-connected to $x_j$ w.r.t. $\varepsilon$ and $MinPts$ (Connectivity)
\end{dfn}

\begin{dfn}{Noise:}\label{def:db6}
Let $C_1,...,C_{\nclust}$ be the $\nclust$ clusters of the given dataset $X$ w.r.t. parameters $\varepsilon_i$ and $MinPts_i$, $i=1,...,\nclust$. Then noise is defined as the set of objects in the dataset $X$ that do not belong to any cluster $C_i$, i.e. $noise = \lbrace x_i \in X \vert \forall i: x_i \notin C_i \rbrace$
\end{dfn}

In Definition \ref{def:db2} an object is a core point if it has at least $MinPts$ number of objects within its $\varepsilon$-neighborhood. In the case that no objects in a given dataset are density-reachable then we would obtain $\nclust$ clusters where $\nclust$ denotes the number of core-points in a dataset $X$ for a given $\varepsilon$ and $MinPts$. This means that the number of core points can be considered as an upper bound for the number of emerging clusters for a given $\varepsilon$ and $MinPts$. Further it can be deduced from the core point definition that the region surrounding a core point is \textit{more dense} compared to density-connected objects that do not satisfy $\vert \mathcal{N}_{\varepsilon} (x_j) \vert \geq MinPts$ meaning that they are objects in more \textit{spare} regions.

\section{Inferring the topological structure enhances clusterability}\label{sec:problems}

In this section, we demonstrate that the correct use of manifold learning (here, specifically: UMAP), as motivated by our topological framing, largely avoids several frequently described challenges in cluster analysis. 

A major problem affecting cluster analysis is that clustering often becomes more challenging in high-dimensional datasets. Specifically, the presence of many irrelevant and/or dependent features potentially degrades results \citep{kriegel2009clustering}. However, contrary to widespread ``folk-methodological'' superstitions and some sources like \cite{assent2012clustering}, the well-known result that $L_p$ distances lose their discriminating power in high dimensions \citep[][e.g.]{beyer1999nearest} is entirely irrelevant for well-posed clustering problems: both the original publication and subsequent works like \cite{kriegel2009clustering} and \cite{zimek2015blind} show that the conditions for this result do not apply if the data is distributed in well separable clusters. In particular, this means that DBSCAN, being based on pairwise distance information, can easily detect clusters in high-dimensional datasets. 

Nevertheless, there are other problems specific to density-based clustering, and DBSCAN in particular, among which finding a suitable density level is one of the most important \citep{kriegel2011density, assent2012clustering}. A recent review  \citep{schubert2017dbscan}, outlined some heuristic rules for specifying $\varepsilon$ for DBSCAN, but domain knowledge should mostly determine such decisions.
More importantly, density-based clustering is likely to fail for clusters with varying densities. In such cases, a single global density level -- for example, specified via $\varepsilon$ in DBSCAN -- cannot delineate cluster boundaries successfully \citep{kriegel2011density}.  

In addition to these well-known issues, we outline another more subtle, less well-known aspect: 
not only does the difficulty of a clustering problem not necessarily increase for high-dimensional $X$, but clusters may even become easier to detect in higher dimensional (embedding) spaces.

\subsection{Enhancing clusterability of DBSCAN with UMAP}\label{sec:init-exp}

The four example datasets we consider here illustrate the following three points: (1) Density-based clustering works in some but not all high-dimensional settings. (2) Perfect performance may not be achievable even for extensive parameter grid searches, and suitable $\varepsilon$ values are highly problem-specific. (3) Most importantly, manifold learning can considerably enhance clustering both by improving performance and by reducing parameter sensitivity of DBSCAN to the extent that it becomes almost tuning-free. 

The datasets we consider here consist of three clusters sampled from three multivariate Gaussian distributions with different mean vectors. In the first two examples, denoted by $\eone$ and $\etwo$, the covariance matrix for all three Gaussians is the identity matrix, inducing clusters of similar densities. In the latter two examples, $\ethree$ and $\efour$, the covariance matrices differ, inducing clusters of different density. In addition, we consider problems with very different dimensionalities. Observations in setting $\eone$ are sampled from 100-dimensional Gaussians, while observations in setting $\etwo$ are sampled from 1000-dimensional Gaussians. In contrast, observations for $\ethree$ and $\efour$ are sampled from 3-dimensional Gaussians. For $\efour$, an additional 1000 features that are irrelevant for cluster membership are sampled independently and uniformly from $[0, 1]$. 
For each setting, we sample 500 observations from each of the three clusters, i.e. each example dataset consists of 1500 observations in total. The complete specifications of the examples are given in Table \ref{tab:init-exp}.

\begin{table}[h]
    \begin{center}
    %\begin{minipage}{174pt}
    \caption{Specifications of the settings $\eone$, $\etwo$, $\ethree$, and $\efour$. In setting $\efour$ clusters are defined by means of $p = 3$ dimensional Gaussians, yet an additional $1000$ irrelevant features are sampled uniformly from $[0, 1]$, leading to a total dimensionality of $1003$.}
    \label{tab:init-exp}
    \begin{tabular}{@{}llll@{}}
        \toprule
        Setting & p & Means & Variances \\ 
        \midrule
        $\eone$   & $100$  & $\bm{\mu}_i \in \{\textbf{0, 0.5, 1}\}$ & $\sigma_i = 1$ \\
        $\etwo$   & $1000$ & $\bm{\mu}_i \in \{\textbf{0, 0.5, 1}\}$ & $\sigma_i = 1$ \\
        $\ethree$ & $3$    & $\bm{\mu}_i \in \{\textbf{0, 3, 7}\}$ & $\sigma_i \in \{0.1, 1, 3\}$ \\
        $\efour$  & $3$ & $\bm{\mu}_i \in \{\textbf{0, 3, 7}\}$ & $\sigma_i \in \{0.1, 1, 3\}$ \\
        \botrule
    \end{tabular}
    %\end{minipage}
    \end{center}
\end{table}

\begin{figure}[H]
    \centering
    \includegraphics[width=\textwidth]{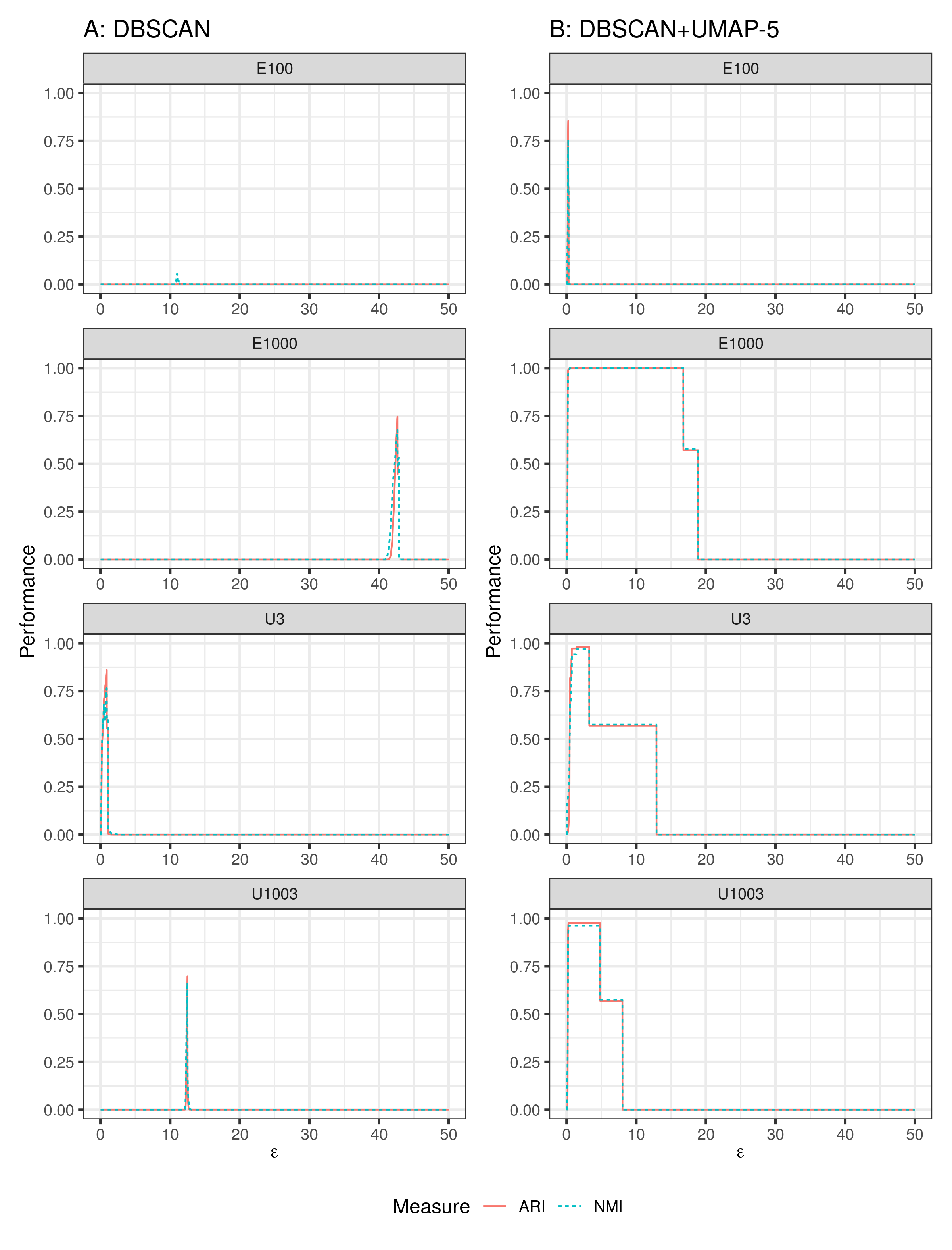}
    \caption{ARI and NMI as a function of $\varepsilon$ for the synthetic settings $\eone, \etwo, \ethree, \efour$. First column: DBSCAN directly applied to the data. Second column: DBSCAN applied to a 2D UMAP embedding with $\kumap = 5$. Clusters sampled from multivariate Gaussian distributions (see Table \ref{tab:init-exp} for specifications). For setting $\efour$, additional $1000$ irrelevant variables are sampled uniformly from $[0, 1]$. DBSCAN computed for $\varepsilon \in [0.01, 50]$, step size: $0.01$; $minPts = 5$.}
    \label{fig:triclust-examples}
\end{figure}

Figure \ref{fig:triclust-examples} shows the Adjusted Rand Index (ARI) \cite[Eq. 5]{hubert1985comparing} and the Normalized Mutual Information (NMI) with maximum normalization \cite[Tab. 2]{vinh2010information} for different $\varepsilon$ values obtained by either applying DBSCAN directly to the observed data or to their 2D UMAP embeddings. Both measures compare two data partitions and return a numeric value quantifying the agreement. While the NMI strictly ranges between $[0, 1]$ (with a value of $1$ indicating perfect concordance), the ARI is $0$ only if the Rand Index exactly matches its expected value under the null hypothesis that the partitions are generated randomly from a hypergeometric distribution \citep{hubert1985comparing}.

Several aspects need to be emphasized. First of all, the effect of the dimensionality of the dataset on the performance of DBSCAN applied to the original data is complicated (Figure \ref{fig:triclust-examples}, first column (A)). Contrary to preconceived notions, it can be easier to detect clusters in higher dimensions. Figure \ref{fig:triclust-examples} A shows that using only DBSCAN, clusters are more easily detected in the $1000$-dimensional data (2nd row) than in the $100$-dimensional data (1st row, although perfect performance is not achieved by DBSCAN in either of the two. 

The dimension of the Gaussian distributions defining the clusters is the only difference between these two settings. On the other hand, Figure \ref{fig:triclust-examples} A, shows that it can also be the other way round. In the 1003-dimensional dataset with 1000 irrelevant features (4th row), cluster performance is much lower than in the corresponding 3-dimensional dataset with only 3 relevant variables (3rd row). Again, perfect cluster performance is not achieved by DBSCAN alone. Note that settings $\ethree$ and $\efour$ define clusters with varying densities, so DBSCAN is expected not to provide a perfect result.

Secondly, finding a suitable value of $\varepsilon$ is very challenging using DBSCAN alone. Note that the optimal $\varepsilon_{opt}$ varies between $0.9$ and $42.64$ for these examples. 
Identifying a suitable $\varepsilon$ is even more problematic since the sensible $\varepsilon$-ranges are very small (e.g. see $\efour$). 
In some cases, clustering does not seem feasible at all even with an optimally chosen $\varepsilon$ --  optimal results are very poor for setting $\eone$ with ARI (NMI) $= 0.003 (0.05)$ for $\varepsilon_{opt} = 11.32 (10.98)$. 
Moreover, while $\varepsilon_{opt}$ is not necessarily consistent for datasets with approximately the same dimensionality -- compare $\varepsilon_{opt} = 42.64$ for $\etwo$ to $\varepsilon_{opt} = 12.48$ for $\efour$ -- it can be similar for datasets with very different dimensionality -- compare $\varepsilon_{opt} \sim 11$ for $\eone$ to $\varepsilon_{opt} = 12.48$ for $\efour$. 

Finally, the crucial point we want to highlight with these examples is that inferring the topological structure before clustering by applying DBSCAN on UMAP embeddings instead of directly to the data makes all these issues (almost) completely disappear (see Figure \ref{fig:triclust-examples} B). First of all, clustering performance is increased in all four examples; in three it even leads to perfect performances. But not only is performance increased, but UMAP also dramatically reduces the complexity of finding a suitable $\varepsilon$. In all considered cases the sensible $\varepsilon$-ranges start near zero, rapidly reach the optimal value, and remains optimal over a wide range of $\varepsilon$-values in three of the four examples. Note that we do not tune UMAP at all -- we simply set $\kumap = 5$ and leave all other settings at their default values.

We emphasize that perfect performance is obtained for large swaths of the $\varepsilon$-range we consider for the two high-dimensional examples. This suggests that the crucial issue in clustering is not the nominal dimension of the dataset or whether it contains irrelevant features, but rather how separable the clusters are in their ambient space, which is usually simply the $p$-dimensional Euclidean space spanned/defined by the dimensions/features of the data, while the approach taken here attempts to cluster observations after projecting them into a space that is optimized for separability.

In summary, applying DBSCAN on UMAP embeddings not only improved performance considerably, but it also reduced the sensitivity of DBSCAN w.r.t. $\varepsilon$. In particular, suitable $\varepsilon$-ranges started near zero for all considered examples. Our experiments described in section \ref{sec:experiments} show that this holds for complex real data such as fashion MNIST \citep{xiao2017fmnist} as well, where applying DBSCAN on UMAP embeddings not only dramatically improved DBSCAN's performance but even outperformed the recently proposed ClusterGAN \citep{mukherjee2019clustergan} method. In the next subsection, we examine the technical aspects that explain this behavior in a simple toy example.

\subsection{Reasons for improved clusterability}\label{sec:problems:toy}
This section lays out possible reasons for the observed improvements w.r.t clusterability with a detailed analysis of  
the underlying technical mechanisms in a simple toy example. Consider the following distance matrix between six objects:

\begin{equation}\label{eq:uc1}
  \left(\begin{array}{cccccc}
  \textcolor{green}{0} & \textcolor{green}{0.6} & \textcolor{green}{0.7} & 1.3 & 1.2 & 1.5 \\
  \textcolor{green}{0.6} & \textcolor{green}{0} & \textcolor{green}{0.5} & \textcolor{orange}{0.75} & 1.6 & 1.3 \\
  \textcolor{green}{0.7} & \textcolor{green}{0.5} & \textcolor{green}{0} & 1.4 & 1.3 & 1.1 \\
  1.3 & \textcolor{orange}{0.75} & 1.4 & \textcolor{cyan}{0} & \textcolor{cyan}{0.7} & \textcolor{cyan}{0.75} \\
  1.2 & 1.6 & 1.3 & \textcolor{cyan}{0.7} & \textcolor{cyan}{0} & \textcolor{cyan}{0.75} \\
  1.5 & 1.3 & 1.1 & \textcolor{cyan}{0.75} & \textcolor{cyan}{0.75} & \textcolor{cyan}{0} \\
  \end{array}\right)
\end{equation}

Inspecting this distance matrix reveals two clusters of objects, shown here in green and cyan. We set DBSCAN's core point condition parameter to $minPts = 2$. Note that the object itself is not considered part of its $\epsilon$-neighborhood. We set $\varepsilon = 0.75$, so that every object whose row (or column) in the distance matrix contains at least two entries $\leq 0.75$ is considered a ``core point''. Since two objects from the different clusters have a distance of exactly $0.75$ (orange entries), all objects are part of a single \textit{connected component}, and the two dense regions are subsumed into a single large cluster for $\varepsilon = 0.75$, as can be seen in the matrix below: 

\begin{equation}\label{eq:uc2}
  \left(\begin{array}{cccccc}
  \textcolor{green}{0} & \textcolor{green}{0.6} & \textcolor{green}{0.7} & 1.3 & 1.2 & 1.5 \\
   \textcolor{green}{0.6} &  \textcolor{green}{0} & \textcolor{green}{0.5} & \textcolor{green}{0.75} & 1.6 & 1.3 \\
   \textcolor{green}{0.7} &  \textcolor{green}{0.5} & \textcolor{green}{0} & 1.4 & 1.3 & 1.1 \\
  1.3 &  \textcolor{green}{0.75} & 1.4 & \textcolor{green}{0} &  \textcolor{green}{0.7} &  \textcolor{green}{0.75} \\
  1.2 & 1.6 & 1.3 &  \textcolor{green}{0.7} & \textcolor{green}{0} &  \textcolor{green}{0.75} \\
  1.5 & 1.3 & 1.1 & \textcolor{green}{0.75} & \textcolor{green}{0.75} & \textcolor{green}{0} \\
  \end{array}\right)
\end{equation}

To avoid this collapsed solution, one could try to reduce the $\varepsilon$ parameter to e.g.  $\varepsilon = 0.74$. 
However, as a consequence, now all the objects in the second (cyan) cluster become ``noise'': They no longer satisfy the ``core point'' condition for $minPts = 2$, since at most one distance in each of their rows is $\leq 0.74$. This means only one cluster (top left, green) is detected, as can be seen in the following matrix:

\begin{equation}\label{eq:uc3}
  \left(\begin{array}{cccccc}
  \textcolor{green}{0} & \textcolor{green}{0.6} & \textcolor{green}{0.7} & 1.3 & 1.2 & 1.5 \\
  \textcolor{green}{0.6} & \textcolor{green}{0} & \textcolor{green}{0.5} &  0.75  & 1.6 & 1.3 \\
  \textcolor{green}{0.7} & \textcolor{green}{0.5} & \textcolor{green}{0} & 1.4 & 1.3 & 1.1 \\
  1.3 &  0.75  & 1.4 & 0 & 0.7 & 0.75 \\
  1.2 & 1.6 & 1.3 & 0.7 & 0 & 0.75 \\
  1.5 & 1.3 & 1.1 & 0.75 & 0.75 & 0 \\
  \end{array}\right)
\end{equation}

From this first example, we conclude 1) that there may be cases where even a single object may \textit{connect} two clusters, yielding a single collapsed cluster and 2) that the sensitivity of clustering solutions to hyperparameter settings is large: A small change of the $\varepsilon$-parameter by only $0.01$ led to a fundamentally different solution.

Thus, we should look for improvements that (i) reduce the sensitivity of results towards the parameter settings and (ii) increase the separability of the data and thereby reduce the susceptibility of DBSCAN to merge multiple poorly separated clusters via interconnecting observations at their respective margins. Sharpening the distinction between dense and sparse regions within the dataset, i.e. increasing separability, improves clusterability. As we will now see, UMAP is able to do exactly that by arranging objects into clusters with fairly constant density within and empty regions in between.

To illustrate this, we consider the representation of the toy example via the fuzzy graph as constructed by UMAP. This reflects the fuzzy simplicial set representation of the data and crucially depends on the number of nearest neighbors $\kumap$. We start with $\kumap = 6$. This leads to a graph with adjacency matrix

\begin{equation}\label{eq:uc8}
  \left(\begin{array}{cccccc}
     0 &  1.0 & 0.95 & 0.29 & 0.53 & 0.25 \\
   1.0 &    0 &  1.0 &  0.9 & 0.19 & 0.30 \\
  0.95 &  1.0 &    0 & 0.24 & 0.45 & 0.58 \\
  0.29 &  0.9 & 0.24 &    0 &  1.0 &  1.0 \\
  0.53 & 0.19 & 0.45 &  1.0 &    0 &  1.0 \\
  0.25 &  0.3 & 0.58 &  1.0 &  1.0 &    0 \\
  \end{array}\right)
\end{equation}

Each cell represents the fuzzy edge weight $v_{ij}$ (Eq. \ref{eq:vij}) connecting two points, so
each value represents the affinity of two observations, not their dissimilarity as in the distance matrices before. As before, the cluster structure is obvious in this representation, with high affinities ($\geq 0.95$) where distances had been low ($\leq 0.75$). The representation learned by UMAP in the graph construction step clearly reflects the cluster structure of the dataset. \\
Note that this fuzzy topological representation by itself already amplifies the cluster structure: if we stopped UMAP at this point and converted the affinities $v_{ij}$ into dissimilarities e.g. via $d_{ij} = 1 - v_{ij},$ $ i \neq j$, DBSCAN with $minPts = 2$ would yield perfect cluster results for $\varepsilon \in [0.01, 0.09]$!\\
Note as well that UMAP's graph layout optimization has not even been performed yet and that the nearest-neighbor parameter $\kumap$ has been set to 6, the largest possible value in this example. Thus, the vast improvement in separability we observe is due only to the way UMAP learns and represents the structure of the data in the fuzzy graph $G$ alone. The improvement can be driven even further both by decreasing the parameter $\kumap$ and by conducting the graph layout optimization. 

First, consider the effect of $\kumap$. In the following, blanks in the matrices denote zero entries. Graph \ref{eq:uc9} shows $G$ for $\kumap = 3$. Clearly, the beneficial effects we noted for $\kumap = 6$ are considerably amplified. 
\begin{equation}\label{eq:uc9}
  \left(\begin{array}{cccccc}
       &  1.0 & 0.83 &      &      &      \\
   1.0 &      &  1.0 & 0.58 &      &      \\
  0.83 &  1.0 &      &      &      &      \\
       & 0.58 &      &      &  1.0 &  1.0 \\
       &      &      &  1.0 &      &  1.0 \\
       &      &      &  1.0 &  1.0 &      \\
  \end{array}\right)
\end{equation}
Almost all $v_{ij}$ become zero (i.e. there is no affinity/similarity between the two points) except for those joined in one of the clusters and the two entries which caused DBSCAN to break. Turning $v_{ij}$ into $d_{ij}$ as above, DBSCAN yields correct clusters for $\varepsilon \in [0.01, 0.42]$. \\
By setting $\kumap = 2$, the smallest possible value due to the local connectivity constraint, we can further distill the cluster structure down to its bare essentials:
\begin{equation}\label{eq:uc10}
  \left(\begin{array}{cccccc}
       &  1.0 &      &      &      &      \\
   1.0 &      &  1.0 &      &      &      \\
       &  1.0 &      &      &      &      \\
       &      &      &      &  1.0 &  1.0 \\
       &      &      &  1.0 &      &      \\
       &      &      &  1.0 &      &      \\
  \end{array}\right)
\end{equation}
Based on this graph, DBSCAN yields correct clusters for $\varepsilon \in [0.01, 0.99]$! Thus, by setting the nearest neighbor parameter of UMAP to a very small value, the cluster separability is dramatically amplified and DBSCAN's sensitivity w.r.t. $\varepsilon$ is significantly reduced.\\
However, the graph layout optimization step has not even been performed yet. This additional step is crucial, in particular for reducing the parameter sensitivity of clustering methods. This is due to the fact $d_{ij} = 1 - v_{ij}$ only converts affinities into dissimilarities. Finding a graph layout via the cross-entropy $C_{UMAP}$ as defined in Eq. \ref{eq:crosse} instead not only converts affinities (indirectly) into dissimilarities but also improves the conversion itself w.r.t. to separability (on top of the separability gained by the graph construction), since the optimization procedure optimizes the graph layout for increased cluster separability. This can be explained as follows:

$C_{UMAP}$ becomes minimal for $v_{ij} = w_{ij}$. For $v_{ij} = 0$, the further away from each other the embedding vectors $y_i$ and $y_j$ are placed, the better, since this will drive $w_{ij}$ towards zero. Considering graphs \ref{eq:uc9} and \ref{eq:uc10}, we see that $v_{ij}$ is zero mostly for observations from different clusters. Minimizing $C_{UMAP}$ thus increases cluster separability in the embedding space by driving objects from different clusters apart. Note that minimizing the cross entropy \lq\lq can be seen as an approximate bound-optimization (or Majorize-Minimize) algorithm [...] implicitly minimizing intra-class distances and maximizing inter-class distances" \cite[p. 3]{boudiaf2020unifying}. The optimization in the graph embedding step of UMAP thus leads to tighter clusters with more white space in between.

The most relevant additional benefit this graph embedding step provides is the large expansion of well-performing $\varepsilon$-ranges for DBSCAN. Since the graph layout optimization uses stochastic gradient descent, the resulting embedding vectors are not deterministic. To account for this randomness, we perform 25 embeddings for each value of $\kumap$ and compute separate averages of the lower and the upper interval boundaries of the $\varepsilon$-ranges yielding optimal cluster performance. On average, the obtained embedding coordinates yield correct clusters for $\kumap = 6$ with $\varepsilon \in [0.83, 1.03]$, for $\kumap = 3$ with $\varepsilon \in [0.70, 6.76]$, and for $\kumap = 2$ with $\varepsilon \in [0.79, 20.94]$. Even the smallest (optimal) $\varepsilon$-ranges we observed over the $3 \times 25$ replications are at least as large as the ones obtained on the fuzzy graph for $k = 6$, and still considerably larger for $k = 3$ and $k = 2$:  $[0.94, 1.03]$, $[0.72, 1.33]$, $[1.16, 4.57]$, respectively. Further analysis of the variability resulting from optimizing embedding vectors via SGD can be found in appendix \ref{sec:app:repl}.

These results indicate how crucial optimizing separability by computing embedding vectors is for clustering performance. Appendix \ref{sec:app:fgraph} confirms its importance on real data.

In these and the following experiments, all of UMAP's other hyperparameters were left to the implementation defaults, in particular $\verb|min_dist| = 0.1$. Additionally adjusting these parameters might further increase separability. However, tuning parameters in an unsupervised setting is a notoriously difficult task and since the results are already convincing by setting $\kumap$ to a small value, we concentrate on the effect of $\kumap$.

In summary, both the graph construction and the graph embedding steps in the UMAP algorithm independently contribute to an increased separability of clusters in a dataset, and their combined effect improves clusterability dramatically.

\section{The price to pay: structures preserved and lost}\label{sec:price}

As we have outlined in the previous sections, UMAP is able to infer and even enhance the topological, i.e. the cluster, structure of a dataset. 
However, these improvements come at a price which will be outlined in this section.

\subsection{Topology vs. geometry}

Beyond topological structure, i.e., mere ``connectedness'', datasets also have geometrical structure -- the shapes of the clusters and how the clusters are positioned relative to each other in the ambient space. 

\begin{figure}[H]
    \centering
    \textbf{A}
    \includegraphics[width=\textwidth, height = 5cm]{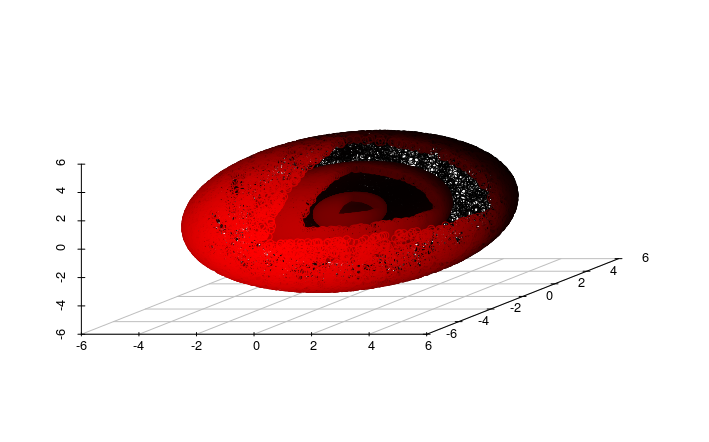}
    \textbf{B}
    \includegraphics[width=\textwidth, height = 5cm]{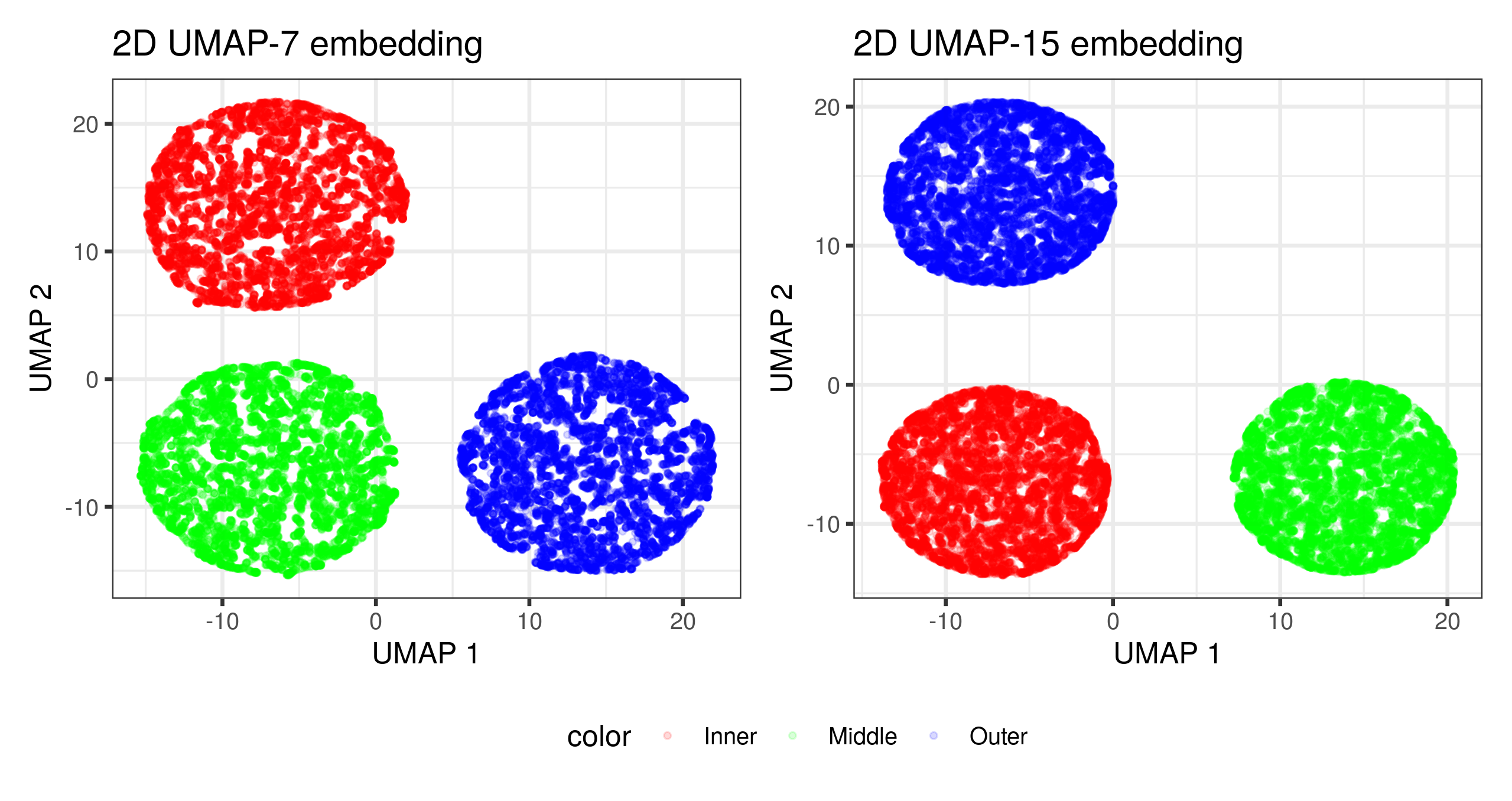}
    \textbf{C}
    \includegraphics[width=\textwidth, height = 5cm]{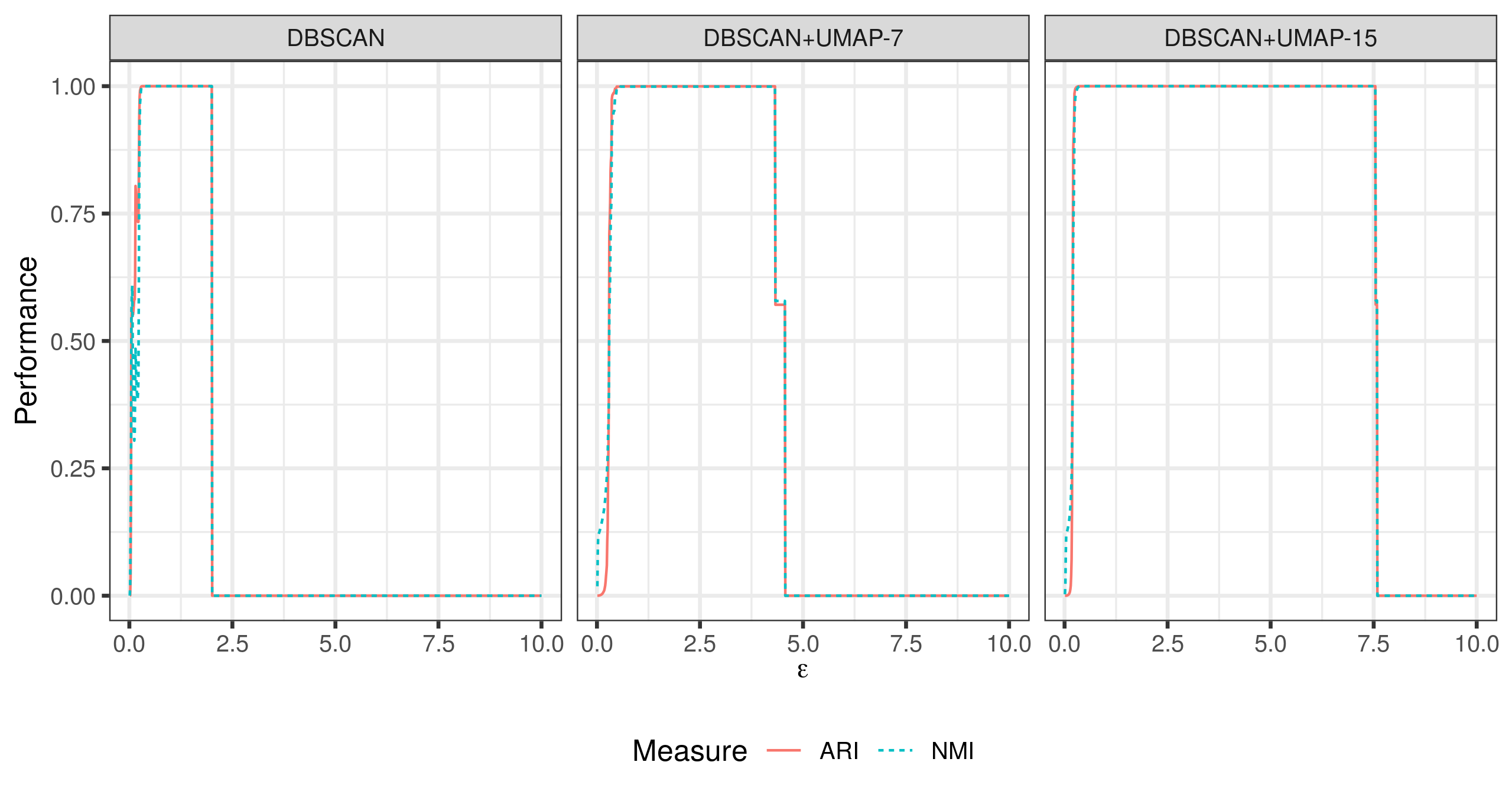}
    \caption{Effects of UMAP: preservation of topological vs. geometrical structure. \textbf{A}: Three nested spheres in 3D ($\nobs = 30000$, part of the data omitted to make the nested structure visible). \textbf{B}: UMAP embeddings for $\kumap = 7$ and $\kumap = 15$. The clusters, i.e. topological structure, is preserved. Geometrical structure is not preserved: Ambient space geometry ("nestedness") is lost; for $\kumap = 7$, less of the spherical/circular shape is preserved. \textbf{C}: Clustering performances for $\varepsilon \in [0, 10]$ (step size: $0.01$, $minPts = 5$) for DBSCAN directly applied to the data (left) and applied to the UMAP embeddings ($\kumap = 7$: middle, $\kumap = 15$: right).} 
    \label{fig:nested-spheres}
\end{figure}

Consider the example of a dataset consisting of three nested spheres embedded in a $3$-dimensional (Euclidean) space (see Figure \ref{fig:nested-spheres} A). What kind of structure does this dataset yield? First of all, from a purely topological perspective, we have three unconnected topological subspaces, i.e. clusters: the three spheres. 

Moreover, from an additional geometrical perspective, we have information on the shape of the individual clusters: they form spheres, i.e. 2-dimensional surfaces. Finally, we have information on the relative position of the clusters to each other within the ambient feature space: the spheres are nested. 

What happens if these data are represented in a 2D UMAP embedding? Since a sphere cannot be isometrically mapped to a 2-dimensional plane, some distortion of the geometric structure will be unavoidable in any 2D embedding. Figure \ref{fig:nested-spheres} B shows that, in fact, most of the geometrical structure is lost in UMAP embeddings: the relative positioning of the clusters diverges from the original data and is not consistent over different embeddings. The effect on the shape of the clusters is less severe. While for $\kumap = 15$ the embeddings are similar to circles, i.e. 2D spheres, for $\kumap = 7$ the general circular shape is retained, yet less uniformly. In contrast, the topological structure of the different clusters is not only preserved in full but even exaggerated -- clusters are much more separated in the embeddings, which is also reflected once again in much wider $\varepsilon$-ranges that yield sensible results (Figure \ref{fig:nested-spheres} C). DBSCAN alone provides perfect clustering performances only over a much smaller $\varepsilon$ range than when applied to these UMAP embeddings.

\begin{figure}[H]
    \centering
    \includegraphics[width=\textwidth, height=17.5cm]{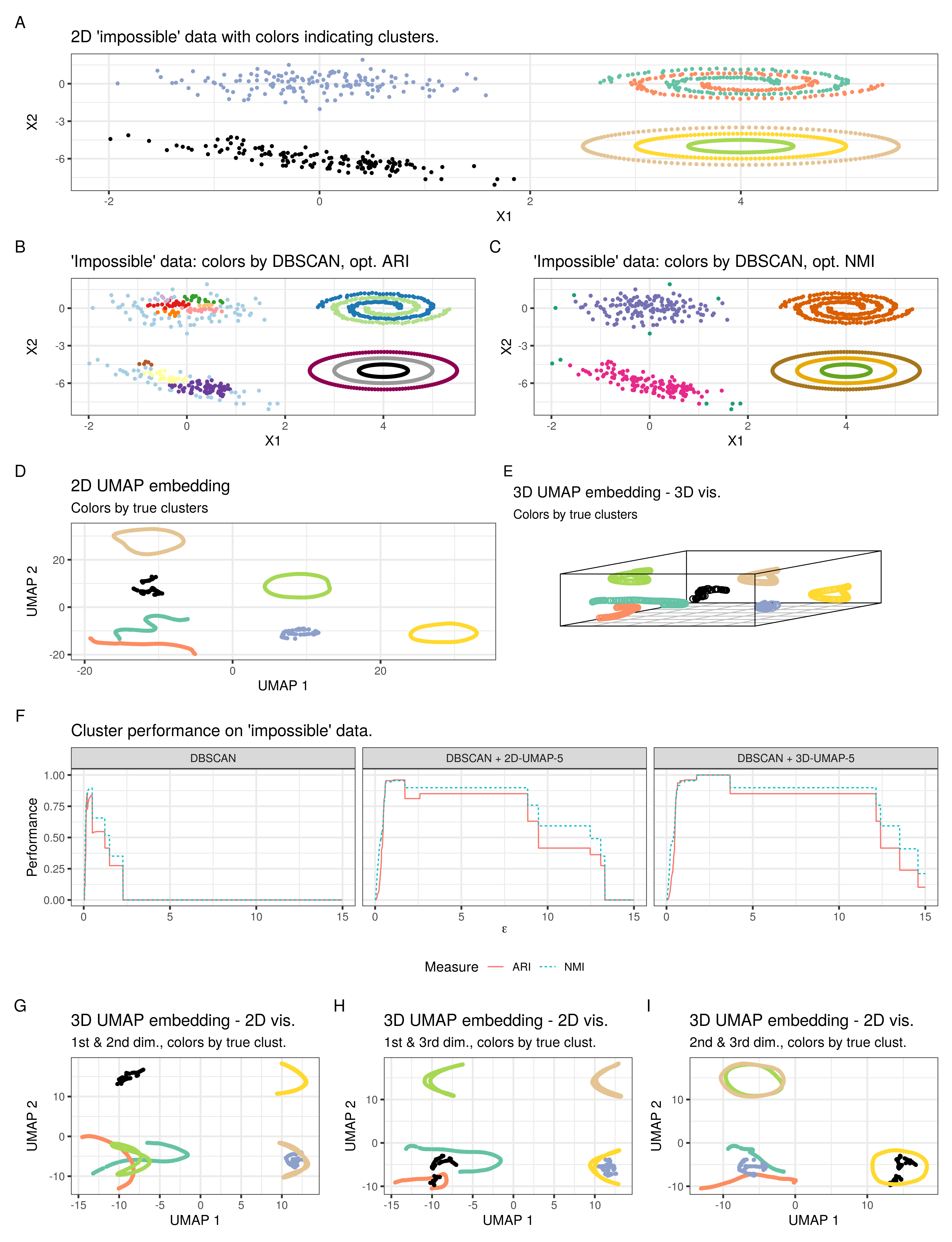}
    \caption{Another example of complex synthetic data and the beneficial effect of \lq\lq dimension inflation". 1st row: the \lq\lq impossible" data with color according to true cluster structure. 2nd row: data colored according to DBSCAN cluster results if applied directly to the data (different optimal $\varepsilon$ values for ARI and NMI).
    3rd row: Visualizations of a 2D and 3D UMAP-5 embedding with colors according to true cluster structure.
    4th row: $\varepsilon$-curves for DBSCAN applied to the data, a 2D UMAP-5, and a 3D UMAP-5 embedding. Last row: 2D visualizations of the 3D UMAP-5 embedding with colors according to true cluster structure. In all settings: DBSCAN computed for $\varepsilon \in [0.01, 15]$, step size: $0.01$; $minPts = 5$.}
    \label{fig:imp-data}
\end{figure}

As a further example, we consider the complex 2D synthetic dataset by \citet{jain2010data}, \lq\lq who suggest that it cannot be solved by a clustering algorithm" \cite[p. 2]{barton2019imp}. 
This \lq\lq impossible" data contains seven clusters with complex structure, see Figure \ref{fig:imp-data} A. The clusters have different densities, are in part non-convex, and are not linearly separable. DBSCAN by itself is not able to detect the full cluster structure and choosing $\varepsilon$ from $[0, 15]$ (step size: $0.01$ $minPts = 5$) based on an optimal ARI value yields a very different cluster result than choosing  $\varepsilon$ based on the optimal NMI value (see Figure \ref{fig:imp-data} B \& C). This challenging example further demonstrates two important points:

First, how successfully UMAP embeddings preserve the connected components (i.e. topological structure) and simultaneously distort geometric structure. In Figure \ref{fig:imp-data} D, we can see that the nested structure of the circles and the entanglement of the spirals are completely lost and that the spirals have been ``unrolled'' in the embedding space, but the different clusters are very clearly separated.

Second, the example illustrates that \lq\lq dimension inflation" via UMAP can have a positive effect on cluster performance. \lq\lq Dimension inflation" means that the data is embedded into a space of higher dimensionality than the observed data. Although this is uncommon and we are not aware of any work where this has been investigated before, there are no restrictions that prevent UMAP from being used in this way. Consider Figure \ref{fig:imp-data} F, which shows ARI- and NMI-curves obtained with DBSCAN applied (1) to the data, (2) a 2D UMAP-5, and (3) a 3D UMAP-5 embedding. Although the 2D UMAP-5 embedding already improves performance and strongly reduces parameter sensitivity, it does not yield a perfect solution. In the 2D embedding (Fig. \ref{fig:imp-data} D), the two spirals are very close to each other, with a gap between them that is smaller than the gap appearing within the black cluster. 

However, the \textit{three} dimensional UMAP-5 embedding not only further reduces parameter sensitivity but also allows for perfect cluster performances. A 3D visualization of this embedding is depicted in Figure \ref{fig:imp-data} E, but note that a static 3D visualization does not make the improved separability visible very well. Figures \ref{fig:imp-data} G-I show all pairwise plots of the three embedding dimensions of the 3D UMAP embedding, even though none of these 2D projections reflects the cluster structure well. We recommend basing exploratory analysis on 3D embeddings as they are more likely to yield good results in complex data than 2D embeddings and still allow for very reasonable visualizations with dynamic plotting tools.

\subsection{Outliers and noise points}

\begin{figure}[H]
    \centering
    \includegraphics[width=\textwidth]{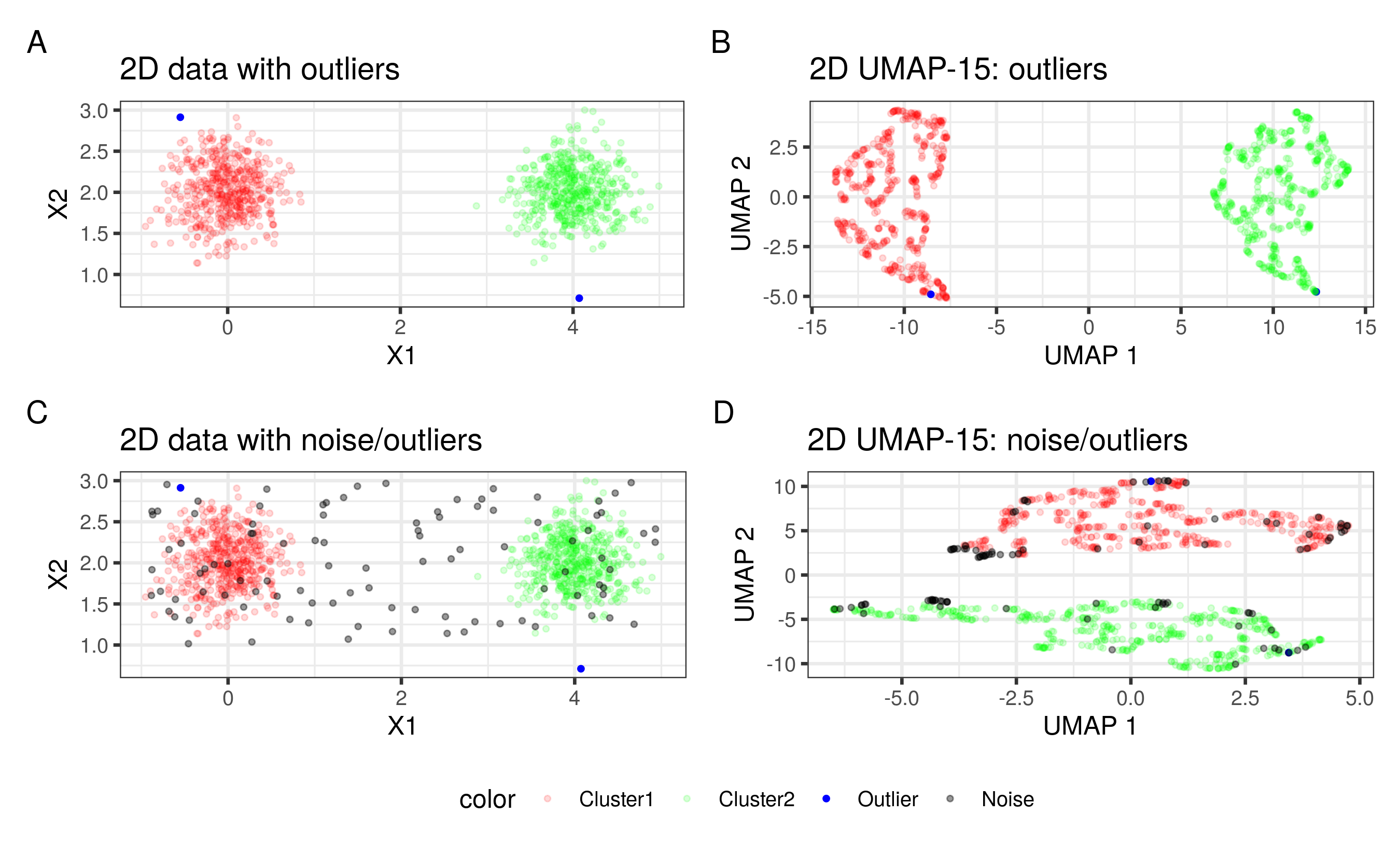}
    \caption{Effect of UMAP on data with outliers and noise points. First column: 2D datasets with two clusters and two outliers (\textbf{A}) and two outliers and noise points (\textbf{C}). Second column: UMAP embeddings with $\kumap = 15$ (\textbf{B} \& \textbf{D}, respectively). The cluster structure is preserved. Outliers and noise points are forced into the clusters.} 
    \label{fig:outlier-and-noise}
\end{figure}

Outliers are another important property of a dataset, but their distinctiveness and relative isolation is unlikely to be preserved in their UMAP embeddings. Consider Figure \ref{fig:outlier-and-noise} A \& C, which shows two 2D datasets with two clusters and, firstly, with two outliers (in blue) on the left-hand side, and, secondly, with additional, uniformly distributed noise points (in grey) on the right-hand side. Corresponding UMAP embeddings for $\kumap = 15$ are depicted in Figure \ref{fig:outlier-and-noise} B \& D. Although the cluster structure is preserved, in both cases the outliers are no longer detectable as such (note that no dimension reduction has taken place). Similarly  for noise points, which are embedded into proximal clusters and then no longer detectable as noise. 

It has recently been shown for functional data that outlyingness can be seen as a metric structure of a dataset \citep{herrmann2021geometric}. Since UMAP does not preserve metric structure (i.e. distances) but connected components, the loss of the outlier structure is not surprising. Moreover, note that UMAP's local connectivity constraint, which ensures that each point is at least connected to its nearest neighbor, may render it generally impossible to preserve outlier structure in UMAP embeddings. Applying outlier detection methods in an additional preprocessing step before computing UMAP embeddings may solve this issue.

\subsection{Overlapping and diffuse clusters}

Clusters with considerable overlap or diffuse boundaries that result in a large likelihood of \lq\lq bridge" points between nominally distinct clusters are especially challenging for most clustering algorithms. 

\begin{figure}[ht]
    \centering
    \includegraphics[width=\textwidth]{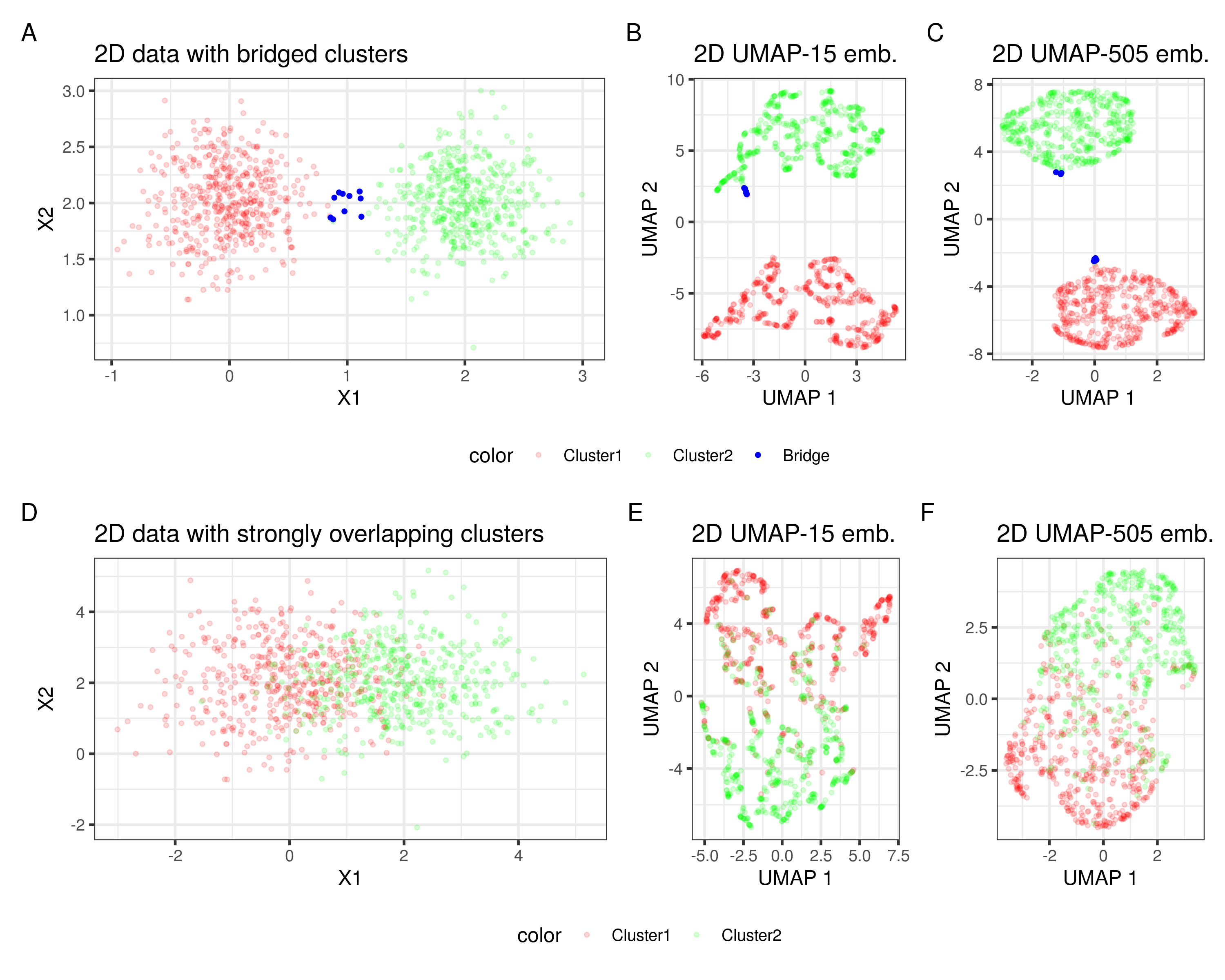}
    \caption{Effect of UMAP on data with connected components. Upper row:  2D data with two bridged clusters. Lower row: 2D dataset with two strongly overlapping clusters. \textbf{A} \& \textbf{D}: data. \textbf{B, C, E, F}: UMAP embeddings with $\kumap = 15$ and $\kumap = 505$, respectively.
    UMAP breaks the bridged components up into two clusters but does not break up the strongly overlapping components.} 
    \label{fig:bridges}
\end{figure}

First of all, consider Figure \ref{fig:bridges} A, which shows a 2D dataset consisting of two clusters that are connected by a small \lq\lq bridge" of points (blue). From a purely topological perspective, we have a single connected topological subspace. A 2D UMAP representation, however, breaks the connected components apart, see Figure \ref{fig:bridges} B \& C. Note, that this holds for a small value of $\kumap = 15$ as well as for a very large value of $\kumap = 505$.
Another issue concerns clusters with substantial overlap, which are often modeled as diffuse components of a Gaussian mixture \citep{rasmussen1999infinite}. In such cases, UMAP and similar manifold learning methods are unlikely to improve clustering performance. Consider Figure \ref{fig:bridges} D. It shows a 2D dataset with two clusters following 2-dimensional Gaussian distributions with mean vectors $(0, 2)'$ and $(2, 2)'$ and unit covariance matrix. Note that in both embeddings (Figure \ref{fig:bridges} E \& F) the clusters are not clearly separable, and the less so the larger UMAP's locality parameter $\kumap$ is chosen.

For strongly overlapping clusters, it is questionable to even consider such settings as (\lq\lq pure") clustering tasks. From a topological perspective, such settings cannot be considered a well-posed clustering problem as there are no separable components in the data. However, in the presence of bridges, it seems reasonable to consider the dataset as consisting of two clusters. Whether overlapping clusters should be merged or considered separate must surely be answered w.r.t. the specific domain. Practitioners should be aware of how UMAP tends to behave in such settings: it typically breaks ``bridges'' apart and merges highly overlapping clusters.

\subsection{Quantitative analysis of further synthetic data} 

In addition to the qualitative analyses of these toy datasets we investigate further examples quantitatively in this paragraph. The datasets under consideration are those from the Fundamental Clustering Problem Suite (FCPS) \citep{ultsch2005clustering}. These datasets are constructed such that they reflect specific clustering problems. Table \ref{tab:fcps-dat} shows key characteristics of these datasets and the problems they present. More details including visualizations can be found in the corresponding papers \citep{thrun2020clustering, ultsch2020fundamental}. 
\begin{table}[h]
    \begin{center}
    %\begin{minipage}{174pt}
    \caption{Characteristics of the FCPS datasets: the number of clusters $\nclust$, the number of observations $\nobs$, the number of features (dimensionality) $p$, and the problem as specified in corresponding papers \citep{thrun2020clustering, ultsch2020fundamental}.}\label{tab:fcps-dat}
    \begin{tabular}{@{}lllll@{}}
        \toprule
        Name & $\nclust$ & $\nobs$ & $p$ & Problem \\
        \midrule
        Hepta       & 7 & 212  & 3 & different variances \\
        LSun        & 3 & 400  & 3 & different variances \& inter cluster distances \\
        Tetra       & 4 & 400  & 3 & almost touching clusters \\
        Chainlink   & 2 & 1000 & 3 & not linearly separable \\
        Atom        & 2 & 800  & 3 & different variances \& not linearly separable \\
        EngyTime    & 2 & 4096 & 2 & Gaussian mixture \\
        Target      & 6 & 770  & 2 & outliers \\
        TwoDiamonds & 2 & 800  & 2 & cluster borders defined by density \\
        Wingnut     & 2 & 1070 & 2 & density vs. distance \\
        Golfball    & 1 & 4002 & 3 & no clusters at all \\
        \botrule
    \end{tabular}
    %\end{minipage}
    \end{center}
\end{table}

The results of applying DBSCAN directly to the data and on 2D UMAP embeddings with $\kumap = 10$ are shown in Table \ref{tab:fcps-res}. Depicted are the highest achievable ARI and NMI values by approach and dataset as well as the $\varepsilon$-range $\varepsilon_{[ARI > 0]}$ for which ARI is greater than zero.

The results show that DBSCAN alone already yields perfect clustering performance for the datasets Hepta, Lsun, Chainlink, Atom, Target, WingNut, and GolfBall. However, note that UMAP clearly reduces $\varepsilon$ sensitivity (much wider $\varepsilon$-range), i.e. it increases clusterability for Hepta, Lsun, Chainlink, Atom, Target. 

On the datasets Tetra and TwoDiamonds, DBSCAN does not perform perfectly. These datasets represent problems (specified as \lq\lq almost touching clusters" (Tetra) and \lq\lq cluster borders defined by density" (TwoDiamonds)) with less clearly separable clusters. Consistent with the examples presented in section \ref{sec:init-exp}, inferring the topological structure via UMAP not only drastically reduces $\varepsilon$ sensitivity of DBSCAN, but it also improves clustering performance to (almost) perfect results in these examples.

In contrast to that, inferring the relevant structure is not possible with UMAP in the settings EngyTime and Target and thus it does not improve the performance of DBSCAN, it even reduces it. This is consistent with the results of the previous subsections: EngyTime is a setting with clusters that overlap strongly, while the Target data is a setting with six clusters of which four are defined by a few outliers. 

\begin{table}[h]
    \begin{center}
    %\begin{minipage}{174pt}
    \caption{Maximum ARI and NMI and $\varepsilon$ ranges corresponding to ARI $> 0$ for FCPS data.}\label{tab:fcps-res}
    \begin{tabular}{@{}llllllll@{}}
        \toprule
             & \multicolumn{3}{l}{DBSCAN} & \multicolumn{3}{l}{$\quad$ UMAP + DBSCAN} \\
        Data & ARI & NMI & $\varepsilon_{[ARI > 0]}$ & & ARI & NMI & $\varepsilon_{[ARI > 0]}$\\
        \midrule
        Hepta & 1 & 1 & [0.0, 2.3] & & 1 & 1 & [0.1, 19]\\
        Lsun  & 1 & 1 & [0.1, 0.7] & & 1 & 1 & [0.1, 14]\\
        Tetra & 0.91 & 0.85 & [0.2, 0.5] & & 0.99 & 0.99 & [0.1, 7]\\
        Chainlink & 1 & 1 & [0.0, 0.8] & & 1 & 1 & [0.0, 7]\\
        Atom & 1 & 1 & [0.8, 20] & & 1 & 1 & [0.0, 13]\\
        EngyTime & 0.36 & 0.23 & [0.0, 1] & & 0.29 & 0.26 & [0.0, 0.9]\\
        Target & 1 & 0.97 & [0.0, 2.3] & & 0.97 & 0.88 & [0.0, 11]\\
        TwoDiamonds & 0.95 & 0.85 & [0.0, 0.1] & & 1 & 1 & [0.0, 4.7]\\
        WingNut & 1 & 1 & [0.1, 0.3] & & 1 & 1 & [0.0, 8.1]\\
        GolfBall & 1 & NaN & [0.0, 20] & & 1 & NA & [0.0, 20]\\
        \botrule
    \end{tabular}
    %\end{minipage}
    \end{center}
\end{table}

In summary, the synthetic examples investigated in this and the previous section show that inferring the topological structure of a dataset can dramatically improve and simplify clustering: improvement in the sense that cluster detection with DBSCAN is considerably more reliable, and simplification in the sense that finding good parameters for DBSCAN becomes significantly less challenging: the suitable $\varepsilon$-ranges are typically much wider, they consistently start near zero and ARI/NMI quickly reach their optimum in this range, so that a quick and simple coarse grid search over small values of $\varepsilon$ is likely to be successful. 

We emphasize that these conclusions apply to diverse and challenging synthetic data settings that include low-dimensional as well as high-dimensional data, data with equal and unequal cluster densities, data with (many) irrelevant features, clusters of arbitrary shape, and not linearly separable clusters. In the next section, we show that this also holds for several real datasets. \\

\section{Experiments on Real-World Data}\label{sec:experiments}

\begin{figure}[H]
    \centering
    \includegraphics[width=\textwidth]{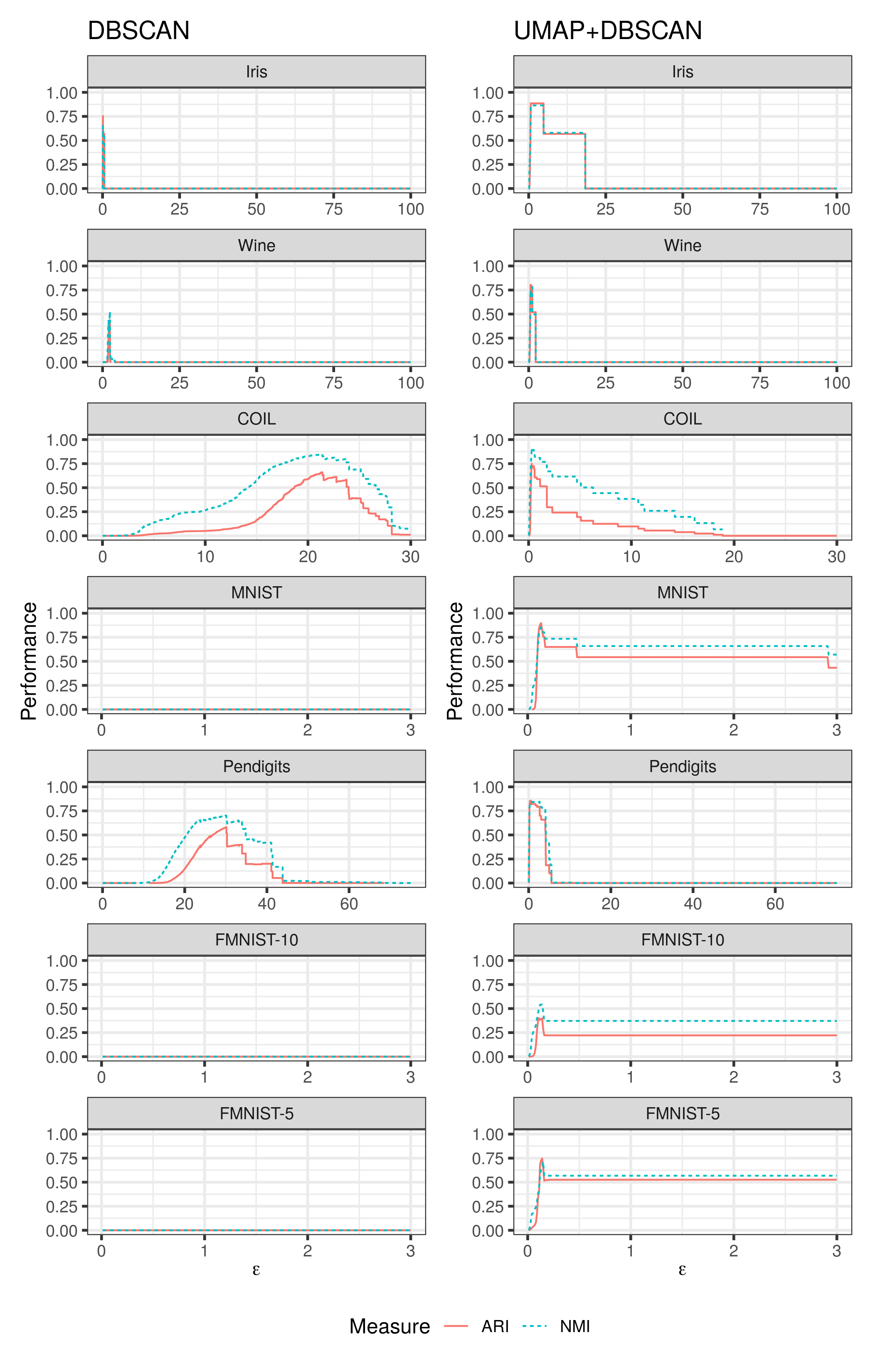}
    \caption{ARI and NMI as functions of $\varepsilon$ for the real datasets. Parameters: $\kumap = 10$ and $d = 3$ (UMAP); $minPts= 5$, $\varepsilon$-step-size = $0.01$ (DBSCAN).} 
    \label{fig:res-real}
\end{figure}

An overview of the real datasets used in this study is given in Table \ref{tab:real-dat}. Since some of these datasets have already been used in other studies, we can investigate not only how the clustering performance of DBSCAN is improved if the topological structure of a dataset is inferred beforehand. We can additionally compare our results to those reported for other clustering methods. The set of datasets includes the well known Iris data \citep{anderson1935irises, fisher1936use}, the Wine data \citep{aeberhard1994wine, forina1998extendible, DuaUCI}, the Pendigits data \citep{alimouglu2001combining, DuaUCI} as well as the COIL \citep{coil20}, MNIST \citep{lecun2010mnist} and fashion MNIST (FMNIST) \citep{xiao2017fmnist} data. Following \cite{mukherjee2019clustergan},  we use two different versions of FMNIST: one with the original ten clusters and a version reduced to five clusters which are pooled from the original ten based on their similarity. The results of applying DBSCAN directly to the datasets and to the embeddings obtained with UMAP are depicted in Figure \ref{fig:res-real} and Table \ref{tab:res-real}.

\begin{table}[h]
    \begin{center}
    %\begin{minipage}{174pt}
    \caption{Characteristics of the real datasets: the number of clusters $\nclust$, the number of observations $\nobs$, and the number of features (dimensionality) $p$. As in the ClusterGAN paper \citep{mukherjee2019clustergan} we investigate two versions of FMNIST: FMNIST-10 and FMNIST-5, the clusters in the latter are:  1: Tshirt/Top, Dress; 2: Trouser; 3: Pullover, Coat, Shirt; 4: Bag; 5: Sandal, Sneaker, Ankle Boot.} \label{tab:real-dat}
    \begin{tabular}{@{}llll@{}}
        \toprule
        Name & $\nclust$ & $\nobs$ & $p$    \\
        \midrule
        Iris      & 3  & 150   & 5          \\
        Wine      & 3  & 176   & 14         \\
        COIL      & 20 & 1440  & 16385      \\
        Pendigits & 10 & 10992 & 17         \\
        MNIST     & 10 & 70000 & 784        \\
        FMNIST-10 & 10 & 70000 & 784        \\
        FMNIST-5  & 5  & 70000 & 784        \\
        \botrule
    \end{tabular}
    %\end{minipage}
    \end{center}
\end{table}

Figure \ref{fig:res-real} shows ARI and NMI as a function of $\varepsilon$ for the different datasets. Table \ref{tab:res-real} details the optimum ARI and NMI achieved within the considered $\varepsilon$-ranges. We inferred the topological structure of the datasets for three different values of $\kumap \in \{5, 10, 15\}$. Note that we did not tune UMAP at all and used \verb|min_dist| $= 0.1$, \verb|n_components| $= 3$ and spectral initialization throughout. Iris and Wine data features were scaled respectively standardized. 

In general, the results show that what has been observed for the synthetic examples also holds for real data. For all considered settings, inferring the topological structure of the dataset via UMAP before applying DBSCAN leads to better clustering performances than applying DBSCAN directly, dramatically so for MNIST and FMNIST. Moreover, it reduces $\varepsilon$ sensitivity of DBSCAN with suitable $\varepsilon$-ranges starting close to zero and with high ($> 0.5$) ARI and NMI values for large parts of the $\varepsilon$-range. 
For DBSCAN directly applied to (F)MNIST, we additionally scanned the $\varepsilon$-range $[0, 100]$ with a step size of $0.1$, but performance did not improve over this extended search grid.

We also investigate the effect of optimizing the separability by constructing embedding vectors instead of using the fuzzy edge weights directly for datasets Iris, Wine, COIL, and Pendigits. Clustering using UMAP's fuzzy graph weights directly performs worse, as expected. For example on the Iris data, computing embedding vectors with UMAP-10 leads to optimal ARI/NMI $= 0.89/0.86$ over an $\varepsilon$-range of $[0.67, 4.82]$ in contrast to $0.88/0.84$ over $[0.6, 0.61]$ if only the fuzzy graph weights of UMAP-10 are used. Both variants still yield better results than applying DBSCAN directly to the data (optimal ARI/NMI $= 0.75/0.67)$. We found similar results for Wine, COIL, and Pendigits, see appendix \ref{sec:app:fgraph}.

\begin{table}[h]
    \begin{center}
    %\begin{minipage}{174pt}
    \caption{Maximum ARI and NMI for the real datasets. DBSCAN directly applied to the data and to 3D UMAP embeddings for $\kumap \in \{5, 10, 15\}$. For the explored $\varepsilon$-ranges, see Fig. \ref{fig:res-real}.}\label{tab:res-real}
    \begin{tabular}{@{}lclccccccc@{}}
        \toprule
        & \multicolumn{2}{c}{DBSCAN} & \multicolumn{2}{c}{DBS+UMAP-5} & \multicolumn{2}{c}{DBS+UMAP-10} & \multicolumn{2}{c}{DBS+UMAP-15} \\ 
        \cline{2-3} \cline{4-5} \cline{6-7} \cline{8-9} 
        & ARI & NMI & ARI & NMI & ARI & NMI & ARI & NMI \\
        \midrule
        Iris      & 0.75 & 0.67 & 0.70 & 0.75 & 0.89 & 0.86 & 0.89 & 0.86 \\
        Wine      & 0.44 & 0.52 & 0.81 & 0.77 & 0.81 & 0.78 & 0.80 & 0.79 \\
        Pendigits & 0.58 & 0.70 & 0.80 & 0.82 & 0.86 & 0.85 & 0.83 & 0.85 \\
        COIL      & 0.66 & 0.85 & 0.82 & 0.93 & 0.75 & 0.91 & 0.70 & 0.88 \\
        MNIST     & 0.00 & 0.00 & 0.69 & 0.70 & 0.90 & 0.85 & 0.87 & 0.85 \\
        FMNIST-10 & 0.00 & 0.00 & 0.41 & 0.59 & 0.40 & 0.54 & 0.38 & 0.54 \\
        FMNIST-5  & 0.00 & 0.00 & 0.60 & 0.62 & 0.75 & 0.71 & 0.63 & 0.63 \\
        \botrule
    \end{tabular}
    %\end{minipage}
    \end{center}
\end{table}

In addition, our results show that the fast, simple and very easily tuneable approach we have proposed leads to comparable or superior clustering performances than recently proposed clustering methods such as ClusterGAN \citep{mukherjee2019clustergan} and SPECTACL(N) \citep{hess2019spectacl} in some settings. Table \ref{tab:comp-real} lists the highest results obtained on the respective datasets in other studies \citep{goebl2014finding, mautz2017towards, mukherjee2019clustergan, hess2019spectacl}. On Pendigits and FMNIST-5, DBSCAN applied to UMAP embeddings performs better than the best-performing methods FOSSCLU and ClusterGAN as reported by \cite{goebl2014finding}, \cite{mautz2017towards}, and \cite{mukherjee2019clustergan}. On MNIST, comparable performance is achieved w.r.t. ClusterGAN and better performance w.r.t. SPECTACL(N). Only for the Wine data and FMNIST-10 are better performance reported for methods FOSSCLU, LDA-k-means, and ClusterGAN. 

\begin{table}[h]
    \begin{center}
    %\begin{minipage}{174pt}
    \caption{Optimal ARI and NMI for some of the real datasets reported in other studies and the methods used. The last two columns show the corresponding optimal performances achieved with DBSCAN \& UMAP.}\label{tab:comp-real}
    \footnotesize
    \begin{tabular}{@{}llllllll@{}}
        \toprule
        Study & Conf. & Data & ARI & NMI & Method(s) & ARI & NMI \\
              &       &      &     &     &           & \multicolumn{2}{l}{(DBS+UMAP)} \\
        \midrule
        \citeauthor{goebl2014finding},   & IEEE & Pendigits & NA   & 0.77 & FOSSCLU     & 0.86 & 0.85 \\
        \citeyear{goebl2014finding}      &      & Wine      & NA   & 0.87 & FOSSCLU     & 0.80 & 0.79 \\
        \citeauthor{mautz2017towards},   & KDD  & Pendigits & NA   & 0.77 & FOSSCLU     & 0.86 & 0.85 \\
        \citeyear{mautz2017towards}      &      & Wine      & NA   & 0.93 & LDA-k-means & 0.80 & 0.79 \\
        \citeauthor{mukherjee2019clustergan},& AAAI & Pendigits & 0.65 & 0.73 & ClusterGAN  & 0.86 & 0.85 \\
        \citeyear{mukherjee2019clustergan}&     & MNIST     & 0.89 & 0.90 & ClusterGAN  & 0.90 & 0.85 \\
                                         &      & FMNIST-10 & 0.50 & 0.64 & ClusterGAN  & 0.41 & 0.59 \\
                                         &      & FMNIST-5  & 0.48 & 0.59 & ClusterGAN, & 0.75 & 0.71 \\
                                         &      &           &      &      & GAN with bp & \\
        \citeauthor{hess2019spectacl}, \citeyear{hess2019spectacl} & AAAI & MNIST     & NA   & 0.76 & SPECTACL(N) & 0.90 & 0.85 \\
        \botrule
    \end{tabular}
    \end{center}
\end{table}

It must be emphasized that these methods also require analysts to pre-specify a fixed number of clusters that are to be found. ClusterGAN's optimal performances reported in Table \ref{tab:comp-real} were achieved only if the true number of clusters was supplied \citep{mukherjee2019clustergan}. The performance on MNIST considerably deteriorated if the number of clusters was not correctly specified. Recall that one of the major advantages of DBSCAN is that it does not require pre-specifying the number of clusters, in contrast to the complexity of specifying and training ClusterGAN. It should be taken into account, first of all, that a suitable network architecture needs to be defined. Note that standard architectures specified elsewhere had to be adapted for ClusterGAN to achieve satisfactory performance. In addition, the various hyperparameters for the GAN, the SGD optimizer, and the generator-discriminator updating require substantial tuning.
Finally, note that our approach works well in settings with both few and many clusters and for both small and large numbers of observations. This is also in contrast to ClusterGAN, which was \lq\lq particularly difficult [... to train ...] with only a few thousand data points" \cite[p. 4616]{mukherjee2019clustergan}.

\section{Discussion}\label{sec:discussion}

In summary, the presented results show that considering clustering from a topological perspective consistently simplified analysis and improved results in a wide range of settings: from a practical perspective, inferring the topological structure of datasets and representing this structure in suitable embedding vectors that are, in some sense, optimized for separability between the different connected components (dramatically) increased clustering performances of DBSCAN, even outperforming a highly complex deep learning-based clustering method, as long as the clusters did not exhibit large overlap. These insights suggest some conceptual conclusions and raise a number of fundamental questions for cluster analysis, which we will discuss in the following.

To begin with, we argue that two \lq\lq perspectives" on cluster analysis should be more strictly distinguished: on the one hand, settings where the aim is to infer the number of connected components in a dataset (the \lq\lq topological perspective"), and on the other hand, settings where clusters may show considerable overlap (in the following the \lq\lq probabilistic perspective"). If the \lq\lq perspective" (implicitly) taken is not clearly specified, the results of cluster analysis can be misleading. For example, in applied, exploratory analyses relevant information may be lost, while in methodological analyses method comparisons can be misleading.
 
Consider a truly unsupervised and exploratory setting (i.e. the true number of clusters is not known and determining it is a crucial part of the problem) in an applied context. From the \lq\lq topological perspective" applying methods that yield a fixed, pre-specified number of clusters is highly questionable in this situation. If the number of clusters is determined a priori for example via domain knowledge, the analysis cannot falsify these a priori assumptions about the data and may hide any unexpected structure. This seems contradictory to the purpose of an exploratory analysis, where the discovery of unexpected structures can yield valuable new insights.
If, on the other hand, approaches such as elbow-plots of cluster quality metrics are used to determine the number of clusters $\nclust$ in a data-driven way, methods inferring and enhancing connected components should be used in the first place.

Another issue concerns the evaluation of competing methods for clustering using datasets with label information.
Label information can be misleading, in particular, if it is (also) used to pre-specify $\nclust$, as the label information may not be consistent with the unconnected components of a dataset. Consider the FMNIST example, where a simple modification of label information -- merging the original 10 into 5 broader categories --  leads to considerably different results. Note that this change of labels was not introduced here, but in \cite{mukherjee2019clustergan}. 
We assume that the performance of ClusterGAN on FMNIST -- as measured based on the original labels -- was not as convincing as for the other datasets. Since it requires no specialized domain knowledge to assess the general similarity of clusters in this dataset containing images of pieces of apparel, a change of labels is easy to do. But while this change did not improve the performance of ClusterGAN in terms of ARI and NMI by much, it considerably improves the performance of DBSCAN + UMAP. In other words: the labels were presumably changed such that they were much more consistent with the actual unconnected components -- i.e. clusters -- in the data. If only the original ten categories of clothing had been considered here, the method comparison would have been misleading, as the different ability of the methods to identify the (un)connected components of the data would have gone unnoticed. The original label information arguably does not reflect the actual cluster structure of the data. This is likely to be the case in many labeled datasets.

On the other hand, consider settings with overlapping clusters. Taking the topological perspective does not make a lot of sense here, as there are no unconnected components if clusters (strongly) overlap, and our investigations showed that it is, in general, questionable that it is possible to infer such cluster structure with methods that aim to infer connected components. In such settings, one should rather take a \lq\lq probabilistic perspective" and assume that the data follow a joint multi-modal probability distribution, i.e., a mixture of probability distributions. Note that this usually implies some kind of domain knowledge from which it makes sense to assume such structure. Many prominent clustering methods such as k-means, Gaussian Mixture models, or approaches based on the EM algorithm are based on this perspective. It has to be emphasized that our experiments on several widely used real-world benchmark datasets showed that an approach based on the topological perspective, which does not use the true number of clusters as a parameter, can perform comparable or even better than methods that do so.

These considerations raise some important questions. First of all, from a rather practical perspective:
Is it fair to compare methods that require $\nclust$ as a parameter with those that do not? How trustworthy is the widely used approach to evaluate clustering methods using labeled data? Is it at all useful to apply non-probabilistic clustering methods on data with assumed strong cluster overlap?

Moreover, from a rather general conceptual perspective:
Can there be methods that work optimally both in settings with large cluster overlap and settings of high separability? As \citet[p. 19]{schubert2017dbscan} state in that regard: 
\begin{quote}
    \lq\lq To get deeper insights into DBSCAN, it would also be necessary to evaluate with respect to utility of the resulting clusters, as our experiments suggest that the datasets used do not yield meaningful clusters. We may thus be benchmarking on the \lq wrong' datasets (but, of course, an algorithm should perform well on any data)."
\end{quote}
This already points to the problem of \lq\lq wrong" datasets, while on the other hand, they state a method should perform well in any setting. In the light of the insights presented here, we would argue that it is very fruitful to investigate the characteristics of settings in which a method or combination of methods works specifically well or even optimally. As outlined, we consider in particular high cluster overlap in contrast to well separable clusters examples of such settings. The underlying principles are fundamentally different (disconnected domains of the clusters vs. connected domains of the clusters) and may require different, maybe even contradictory objectives to be optimized. This is specifically relevant as a dataset may consist of both sorts of (assumed) structures. We think the insights and results presented here support this view.

\section{Conclusion}\label{sec:conclusion}

This work considered cluster analysis from a topological perspective. Our results suggest that the crucial issue in clustering is not the nominal dimension of the dataset or whether it contains many irrelevant features, but rather how separable the clusters are in the ambient observation space they are embedded in.
Extensive experiments on synthetic and real datasets clearly show that focusing on the topological structure of the data can dramatically improve and simplify cluster analysis both in low- and high-dimensional settings. To demonstrate this principle in practice, we used the manifold learning method UMAP to infer the connected components of the datasets and to create embedding vectors optimized for separability, to which we then applied DBSCAN. 

Using synthetic data, we showed that this makes results much more robust to hyperparameters in a diverse set of problems including low-dimensional as well as high-dimensional data, data with equal and unequal cluster densities, data with (many) irrelevant features and clusters of arbitrary, not linearly separable shapes. The parameter sensitivity of DBSCAN is consistently and dramatically reduced, simplifying the search for a suitable $\varepsilon$-value. Moreover, the cluster detection performance of DBSCAN was considerably improved compared to applying it directly to the data. 

Experiments in real data settings corroborated these insights. In addition, our results showed that the simple approach of combining UMAP and DBSCAN can even outperform complex clustering methods SPECTACL and deep-learning-based ClusterGAN on complex image data such as Fashion MNIST. 

All these results were obtained with very little hyperparameter tuning for UMAP. In particular, we always used a small value of the parameter $\kumap$/\verb|n_neighbors| -- $\kumap \in \{5, 10, 15\}$ in most of our experiments -- markedly reducing the complexity of the parameter choice in density-based clustering. All other parameters were set to the default values. Based on a simple toy example we provided a detailed technical explanation of why the choice of a small $\kumap$ is reasonable for the purpose of clustering. 

Finally, we propose a conceptual differentiation of cluster analysis suggested by the topological perspective and the presented results. Specifically, we argue that settings with high cluster overlap in
contrast to well separable clusters should be considered as fundamentally different settings which require different kinds of methods for optimal results,  a distinction usually not made explicit enough.
We also propose that using external label information to evaluate clustering solutions should only be done if these labels actually correspond to the (un)connected components of the data manifold from which observations are sampled. If this is not the case, we would argue that evaluation metrics diverge from what clustering algorithms should properly optimize for -- identifying (un)connected components -- and results will be misleading.

We think these considerations point out important questions to be investigated in future work.  

\section*{Declaration}

\subsection*{Funding}

This work has been funded by the German Federal Ministry of Education and Research (BMBF) under Grant No. 01IS18036A. The authors of this work take full responsibility for its content.

\subsection*{Conflict of Interest}

The authors have no competing interests to declare that are relevant to the content of this article.

\subsection*{Data and code availability}

All code and data to reproduce the results can be found on GitHub: \url{https://github.com/HerrMo/topoclust}.

\begin{appendices}

\section{Embedding variability}\label{sec:app:repl}

\begin{figure}[H]
    \centering
    \includegraphics[width=\textwidth]{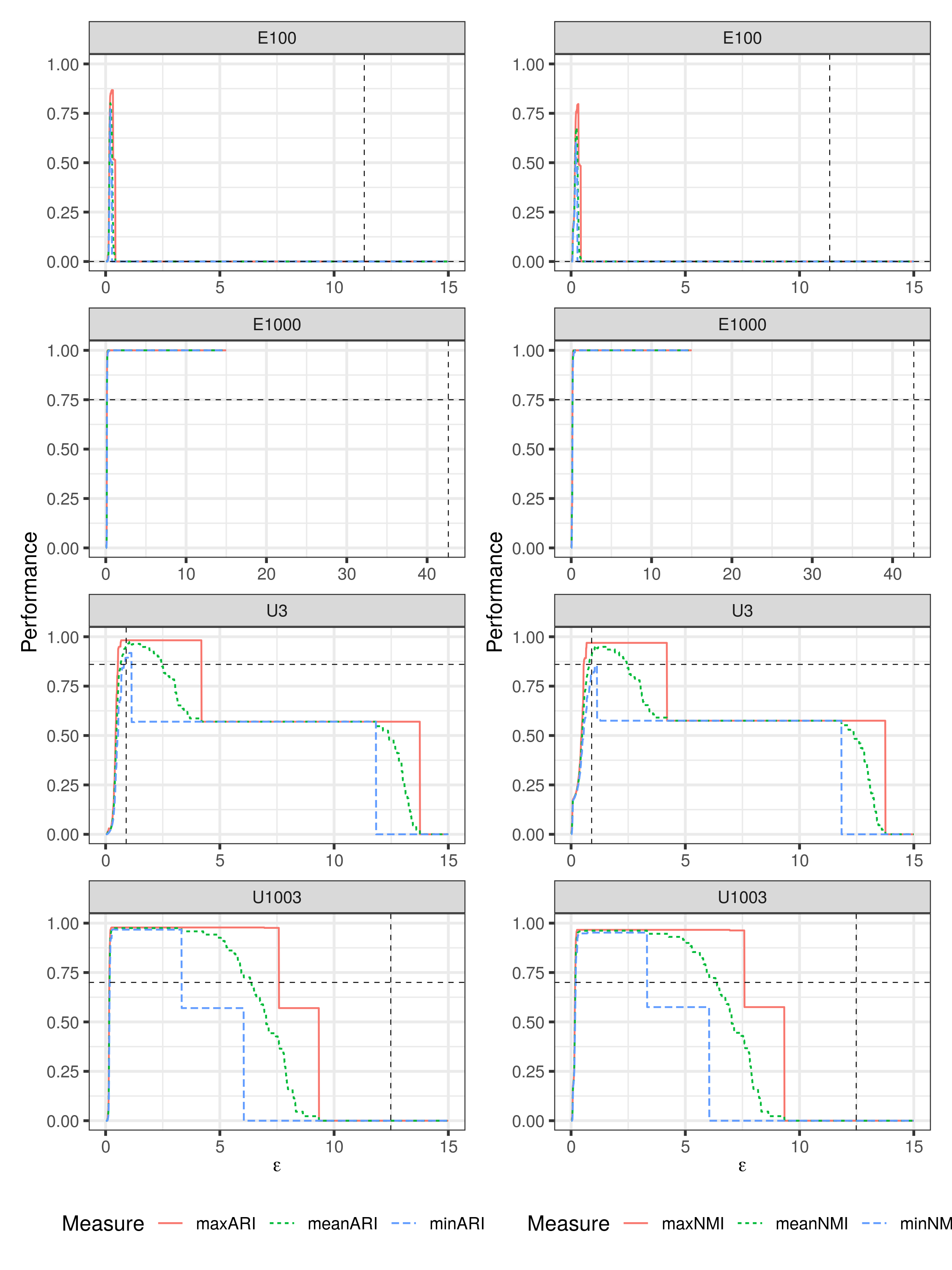}
    \caption{Maximum, mean, and minimum ARI (left column) and NMI (right column) curves summarized over 25 embeddings of the four synthetic settings $\eone, \etwo, \ethree, \efour$. Note, the curves do not reflect a single embedding, but the worst/mean/optimal case over all 25 embeddings for each individual $\varepsilon$-value. The maximum ARI and NMI values obtained by applying DBSCAN directly to the data are shown as a black dashed horizontal line and the corresponding $\varepsilon$-value as a black dashed vertical line. DBSCAN computed for $\varepsilon \in [0.01, 15]$, step size: $0.01$; $minPts = 5$.}
    \label{fig:syn-reps}
\end{figure}

\begin{figure}[H]
    \centering
    \includegraphics[width=\textwidth]{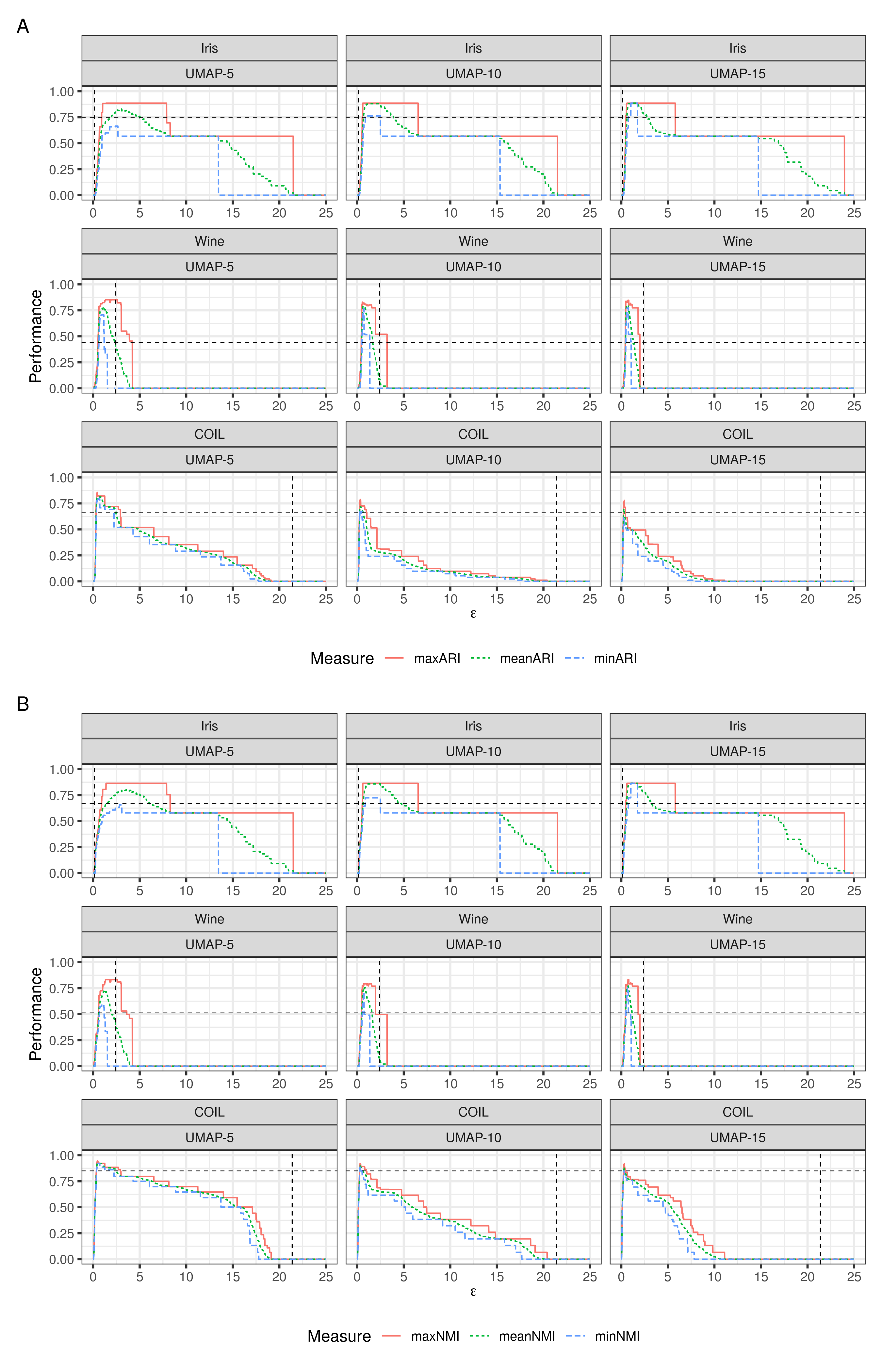}
    \caption{Maximum, mean, and minimum ARI (A) and NMI (B) curves summarized over 25 embeddings of the Iris, Wine, and COIL data. Note, the curves do not reflect a single embedding, but the worst/mean/optimal case over all 25 embeddings for each individual $\varepsilon$-value. The maximum ARI and NMI values obtained by applying DBSCAN directly to the data are shown as a black dashed horizontal line and the corresponding $\varepsilon$-value as a black dashed vertical line. DBSCAN computed for $\varepsilon \in [0.01, 25]$, step size: $0.01$; $minPts = 5$.}
    \label{fig:real-reps}
\end{figure}

In section \ref{sec:problems:toy}, we showed that although the computation of embedding vectors induces some variability with respect to the meaningful $\varepsilon$-range, it also leads to considerably improved separability and is therefore crucial from a clustering perspective. Here we provide additional experiments on this which are based on the synthetic settings from section \ref{sec:init-exp} and the three smallest ($\nobs < 10^4$ observations) real datasets Iris, Wine, and COIL. We computed 25 embeddings for each of the datasets (and $\kumap$ values in the case of the real datasets) and corresponding clusterings on $\varepsilon$-grids $[0.01, 15]$ and $[0.01, 25]$, respectively, with a step size of $0.01$. For each $\varepsilon$-value, the individual minimal, mean, and maximal ARI and NMI values are computed over the 25 replications. Figures \ref{fig:syn-reps} and \ref{fig:real-reps} depict the corresponding minimum, mean, and maximum ARI and NMI curves. Note that the curves do not reflect a single embedding, but the worst/mean/best case over all 25 embeddings for each individual $\varepsilon$-value. In addition, the maximum ARI and NMI values obtained by applying DBSCAN directly to the data are shown as a black dashed horizontal line and the corresponding $\varepsilon$-value as a black dashed vertical line.

In summary, the results again show that optimizing embedding vectors induces some variability with respect to the sensible $\varepsilon$-range across different embeddings. However, this variability can be neglected if the main focus is on improving cluster detection. First of all, the variability does not affect the fact that the sensible $\varepsilon$-ranges start near zero and quickly reach the optimal value, which is in stark contrast to DBSCAN directly applied to the data (see the black dashed horizontal and vertical lines, and Fig. \ref{fig:triclust-examples}). In addition, in all settings, the mean ARI and NMI curves are higher on larger parts of the $\varepsilon$-ranges as the maximum ARI and NMI for DBSCAN directly applied to the data. Note that, except for UMAP-5 on Iris and UMAP-15 on COIL, this holds for the minimum curves as well!

\section{Using just the fuzzy graph weights versus using embedding vectors}\label{sec:app:fgraph}

\begin{figure}[H]
    \centering
    \includegraphics[width=\textwidth]{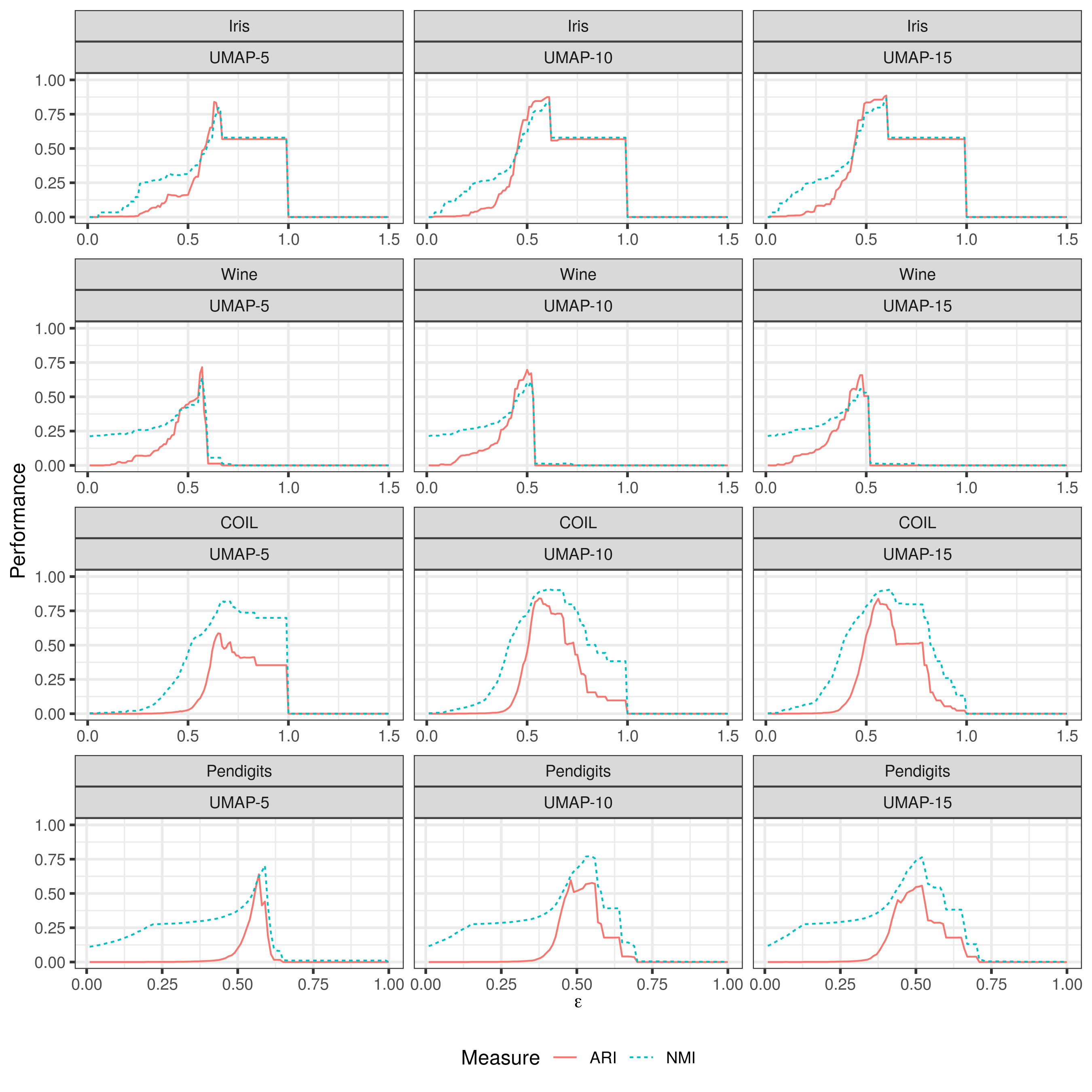}
    \caption{ARI and NMI as a function of $\varepsilon$ for four of the real datasets. Results obtained by applying DBSCAN on the fuzzy graph computed by UMAP (converted into a dissimilarity matrix via $d_{ij} = 1 - v_{ij}$, $i \neq j$, with $v_{ij}$ an edge weight). Embedding vectors optimized for separability have not been constructed. DBSCAN computed for $\varepsilon \in [0.01, 1.5]$, step size: $0.01$; $minPts = 5$.}
    \label{fig:res-fgraph}
\end{figure}

Figure \ref{fig:res-fgraph} shows ARI and NMI as a function of $\varepsilon$ for four of the real datasets. Cluster results were computed using just the fuzzy graph weights, without additionally computing embedding vectors. Converting the graph weights into dissimilarities via $d_{ij} = 1 - v_{ij}$, $i \neq j$, means that the meaningful $\varepsilon$-range is restricted to $[0, 1]$. Moreover, the sensible $\varepsilon$-ranges (yielding optimal or high ARI/NMI values) are smaller than those resulting based on additionally optimized embedding vectors. 

Here we shortly detail this effect for the Wine, COIL, and Pendigits data based on the UMAP-10 results. The Iris data results are exemplarily discussed in section \ref{sec:experiments}.

For the Wine data, only computing the fuzzy graph with UMAP-10 leads to optimal ARI/NMI $= 0.7/0.61$ for a single $\varepsilon = 0.5/0.52$. In contrast, additionally computing optimized embedding vectors leads to ARI/NMI = $0.81/0.78$ for $\varepsilon \in [0.64, 0.69]/[1.11, 1.16]$. Unlike the Iris and Wine data, the optimal ARI/NMI value for the Pendigits and COIL data is only achievable for a single $\varepsilon$-value. Using embedding vectors is nevertheless beneficial. To see this, consider that on Pendigits an ARI/NMI $> 0.6$ can be obtained over $[0.17, 4.04]/[0.16, 4.13]$ with embedding vectors. Only using the fuzzy graph would mean that an ARI $> 0.6$ is not at all achievable and NMI $> 0.6$ only for $\varepsilon \in [0.48, 0.56]$. Similar holds for COIL, with ARI/NMI $> 0.6$ for $\varepsilon \in [0.25, 0.8]/[0.18, 4.07]$ in contrast to $[0.52, 0.68]/[0.45, 0.79]$.

Again, it needs to be emphasized that only using the fuzzy graph still yields better results than applying DBSCAN directly to the data. For example, applying DBSCAN directly to the Wine data yields optimal ARI/NMI $= 0.44/0.52$.

In summary, these investigations also show that computing embedding vectors optimized for separability on top of the fuzzy graph not only reduces parameter sensitivity of the clustering method but can also lead to a better clustering performance due to improved separability.

\section{Real data embedding visualizations}\label{sec:app:emb-vis}

\begin{figure}[H]
    \centering
    \includegraphics[width=\textwidth]{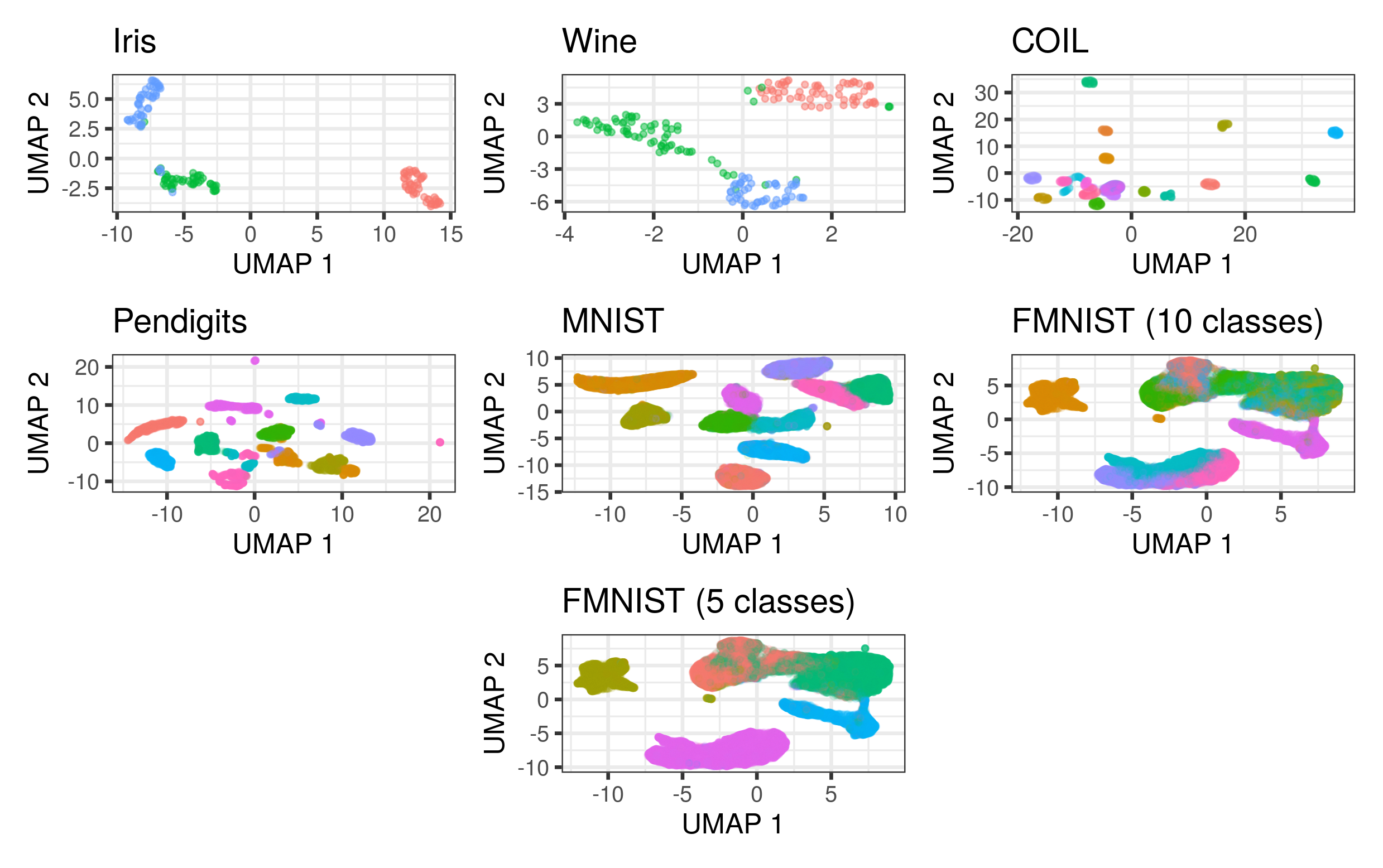}
    \caption{Visualizing 2D UMAP-10 embeddings of the real datasets. Note that an embedding dimension of $d = 2$ was chosen for the purpose of optimal static visualization, in contrast to $d = 3$ used for better cluster detection in the quantitative experiments in section \ref{sec:experiments}.}
    \label{fig:real-emb-vis}
\end{figure}

Figure \ref{fig:real-emb-vis} shows 2D UMAP-10 embeddings of the real datasets under investigation. Colors correspond to the class labels. As can be seen, the inferred connected components clearly agree with the labels for most of the datasets. In FMNIST this holds much better for the 5-label-set. However, note that although 3D embeddings are used in the experiments as they are better suited for cluster detection, they are less well suited for static visualizations (see section \ref{sec:price}). That is why we depict UMAP-10 embeddings optimized in two dimensions (i.e. $d = 2$) here. 

\end{appendices}

\end{document}